\documentclass{article}

\usepackage{arxiv}

\usepackage[utf8]{inputenc}
\usepackage[T1]{fontenc}
\usepackage{times}
\usepackage{hyperref}
\usepackage{url}
\usepackage{booktabs}
\usepackage{tabularx}
\usepackage{longtable}
\usepackage{multirow}
\usepackage{nicefrac}
\usepackage{microtype}
\usepackage{graphicx}
\usepackage{xcolor}
\usepackage{natbib}
\usepackage{doi}

\usepackage{amsmath,amsfonts,bm}

\def\eqref#1{equation~\ref{#1}}
\def\1{\bm{1}}

\def\vs{{\bm{s}}}

\def\vx{{\bm{x}}}
\def\vy{{\bm{y}}}
\def\vz{{\bm{z}}}

\def\mI{{\bm{I}}}

\DeclareMathAlphabet{\mathsfit}{\encodingdefault}{\sfdefault}{m}{sl}
\SetMathAlphabet{\mathsfit}{bold}{\encodingdefault}{\sfdefault}{bx}{n}

\usepackage{cleveref}

\title{Joint Classification and Generation on Tabular Data:\\An Empirical Study}

\renewcommand{\headeright}{}
\renewcommand{\undertitle}{Preprint}
\renewcommand{\shorttitle}{Joint Classification and Generation on Tabular Data}

\author{
  Tim d'Hondt \\
  Eindhoven University of Technology\\
  \texttt{t.dhondt@tue.nl} \\
  \And
  Oleksii Furman \\
  Wroclaw University of Science and Technology\\
  \texttt{oleksii.furman@pwr.edu.pl} \\
  \And
  Damian Kwolek \\
  Wroclaw University of Science and Technology\\
  \texttt{272721@student.pwr.edu.pl} \\
  \And
  Jan Poręba \\
  Wroclaw University of Science and Technology\\
  \texttt{268446@student.pwr.edu.pl} \\
  \And
  Maciej Zieba\\
  Wroclaw University of Science and Technology\\
  Tooploox Sp. z o.o.\\
  \texttt{maciej.zieba@pwr.edu.pl}\\  
  \And
  Mykola Pechenizkiy \\
  Eindhoven University of Technology\\
  \texttt{m.pechenizkiy@tue.nl} \\
}

\hypersetup{
  colorlinks=true, linkcolor=blue!60!black, citecolor=blue!60!black, urlcolor=blue!60!black,
  pdftitle={Joint Classification and Generation on Tabular Data: An Empirical Study of Four Model Families},
  pdfauthor={Tim d'Hondt, Oleksii Furman, Damian Kwolek, Jan Poręba, Maciej Zieba, Mykola Pechenizkiy},
  pdfkeywords={joint modelling, tabular data, energy-based models, energy discrepancy, probabilistic circuits, tabular foundation models, diffusion models, synthetic data, efficiency},
}

\begin{document}
\maketitle

\begin{abstract}
Many applications of machine learning on tabular data need both a classifier and a generator of realistic records. The default is to train two separate models, whereas a single joint model promises both from one set of weights. Several families of tabular deep learning models can be turned into such a joint model, which raises the question of which joint model is preferable at which computational budget.
In this empirical study, we introduce and compare joint models based on four state-of-the-art tabular data modeling families (energy-based models, probabilistic circuits, tabular foundation models and tabular diffusion models) against measured FLOPs and wall-clock time, on 29 classification datasets. Methods are evaluated based on two defined aggregate scores that take into account different aspects of prediction and generation. We then also compare these joint models with every pair of a separate classifier and generator at different computational budgets, to see when a joint model may be a defendible choice against the regular practice of keeping a separate classifier and generator. 
We find that the best joint model is a probabilistic circuit at the smallest budgets and, beyond that, a tabular foundation model that generates column by column using a regressor (TabICL). Compared with pairs of separate models, this study finds a joint model is rarely preferable but often very close to the best pair, without the hassle of training and maintaining two separate models.
\end{abstract}


\section{Introduction}
\label{sec:introduction}

% \begin{figure}[!b]
%   \centering
%   \includegraphics[width=0.85\textwidth]{figures/headline_overall.pdf}
%   \caption{Overall score (geometric mean of the prediction and generation scores) of every joint model against its total FLOPs (fit $+$ predict $+$ sample), on the small OpenML-CC18 tables (left) and the larger UCI-mixed tables (right). Each curve is the best score a method reaches within a given budget over its swept budget hyperparameter. On the small tables one tabular foundation model, TabICL, is best from $1.5\cdot10^{12}$ FLOPs on; on the larger tables a probabilistic circuit comes close to it at a fraction of the cost (\Cref{sec:tradeoff}).}
%   \label{fig:headline}
% \end{figure}

% P1 -- the problem and the promise of one model
In applications of machine learning on tabular data, it is not uncommon to require a generative model of realistic records in addition to a classifier. 
Such a generator can augment or rebalance small and incomplete training sets \citep{cui2024tabular, jager2021benchmark}, support out-of-distribution detection, selective prediction and calibration \citep{liu2020energy, geifman2017selective, grathwohl2019your}, and produce realistic counterfactual explanations and feature attributions \citep{redelmeier2024mcce, aas2021explaining}.
The straightforward way to achieve this is to train two separate models: the best available classifier and the best available generator. 
However a single \emph{joint} model $p(\vx, y)$ could in theory offer both in a single training run and a single set of weights to maintain, which is an attractive proposal in resource-constrained settings. 
% Moreover, hybrid classifier-generator training has been reported to improve the calibration and robustness of the classifier as well \citep{grathwohl2019your, yang2023towards}, and since its density is computed from the same logits as the classifier, it may also be better aligned with the classifier for out-of-distribution detection than the density of a separately trained generator \citep{grathwohl2019your, nalisnick2019deep}.

% TODO: extend this if we add more baselines down the line!
% P2 -- why now: many recent tabular methods can be turned into joint models
The past few years have brought a wave of powerful deep learning methods for tabular data, of which we show some can be turned into such a joint model with little or no modification.
\emph{Tabular foundation models} are transformers pre-trained on millions of synthetic datasets that classify a new table in context and match or outperform tuned gradient-boosted trees on small and medium-sized tables \citep{hollmann2025accurate, qu2025tabicl, kong2026tabfm}. Because any column in a mixed-type table, with numerical and categorical features, can be their prediction target, the chain rule can turn them into generators of whole rows, with the label drawn last from the classifier itself \citep{hollmann2025accurate}.
\emph{Tabular diffusion models} are the state-of-the-art in tabular data generation \citep{zhang2024tabsyn, shi2024tabdiff} and can define a classifier through Bayes' rule, as diffusion classifiers do for images \citep{li2023diffusion}.
\emph{Deep probabilistic circuits} model the joint distribution of mixed-type data natively and have recently been shown to match these generators \citep{scassola2026sobering}.
Finally, \emph{energy-based models} can now be trained on mixed-type data without the costly sampling loop of maximum likelihood \citep{schroder2023energy, schroder2024heat}, which makes Joint Energy-based Models (JEMs) \citep{grathwohl2019your} applicable to tables.

% P3 -- the gap and why it is hard to close
We observe that many of these modern tabular methods have been studied in isolation on either classification or generation, but no work exists that compares their capabilities as joint models on both tasks at the same time, and at what computational cost \citep{vanbreugel2024position}. 
These costs also sit in different places and differ by orders of magnitude: a foundation model is pretrained once but pays at inference with a forward pass over its entire context, a diffusion model is trained on each new dataset for thousands of epochs and needs multiple denoising passes for every class to make one prediction, and an energy-based model pays for sampling during both training and generation.
Moreover, a joint model should not only be compared with other joint models. The natural alternative is to train two separate models, and a joint model is only worth using if it is at least as good as the best such pair at the same cost.
We therefore ask:
\begin{quote}
\emph{Which joint model of tabular data is preferable at which computational budget, and how does it compare with two separate models?}
\end{quote}

% P4 -- what we do
In this empirical study, we aim to answer this question by comparing different deep tabular joint modeling methods on a set of predictive and generative metrics, at different compute budgets, on 29 classification datasets from UCI and OpenML-CC18.  
For every method we sweep the hyperparameter that primarily controls its compute budget and record predictive and generative performance against the amount of FLOPs measured to fit, perform inference and generate samples.
Next to this we also evaluate the performance of state-of-the-art non-joint tabular classification and generation models, and compare the best joint model with the best \emph{pair} of specialist models at each compute budget.

We find that a probabilistic circuit is the best joint model at the smallest budgets and TabICL, a tabular foundation model, at larger budgets. The families differ in characteristic ways: the foundation models are the strongest classifiers but are limited by their context on large tables, the probabilistic circuit and the diffusion models are the best generators on large tables, and the energy-based models classify competitively but generate less realistic samples. Compared with two separate models, a joint model is rarely preferable. It beats every pair only on small tables at medium budgets, where TabICL's own classifier is the best one affordable. Elsewhere a pair is at least as good, and the best pair often keeps the generator of a joint model next to a cheap separate classifier.

% TODO(authors): consider a teaser figure here (the one-vs-two plot answers the title in one image).

Our contributions are:
\begin{itemize}
    \item \textbf{A cost-aware comparison of joint models for tabular data} To our knowledge, this is the first study that compares joint tabular classifiers and generators from four modern deep learning model families, and the first to ask whether a single joint model beats the best pair of separate models at a certain compute budget. We measure for every method the FLOPs used in fitting, prediction and sampling by counting every floating-point operation they execute. We use an aggregated score for prediction and generation of multiple metrics that are all scaled between a trivial floor and real data.
    \item \textbf{New joint tabular models from published components} We define joint models of mixed-type tables that, to our knowledge, have not been evaluated before:
    JEMs \citep{grathwohl2019your} whose marginal energy is trained with energy discrepancy and the mixed-type perturbation kernel of \citet{schroder2024heat} (JEM-ED), or with SGLD and exact Gibbs updates of categorical features (JEM-SGLD). We also evaluate diffusion classifiers \citep{li2023diffusion}, built on the tabular diffusion models TabDiff and TabSyn. Finally, we use column-by-column generators for the foundation models TabPFN and TabICL, following \citet{hollmann2025accurate} and the TabICL package \citep{qu2025tabicl}.
    We give the first quantitative evaluation of each as both a classifier and a generator. 
\end{itemize}

\section{Background}
\label{sec:background}

We consider tabular classification data $(\vx, y)$ with $\vx = (\vx^{\text{num}}, \vx^{\text{cat}}) \in \mathbb{R}^{d_n} \times \prod_{j=1}^{d_c} \{1, \dots, C_j\}$ containing $d_n$ numerical and $d_c$ categorical features, and a label $y \in \{1, \dots, K\}$. We call such data \emph{mixed-type}; purely categorical tables ($d_n = 0$) are included as a special case. This section introduces the four model families we consider in our study in their original form. How we use each of them as a joint model of $(\vx, y)$ is the subject of \Cref{sec:routes}.

\subsection{Energy-based models}
\label{sec:ebm-background}
An energy-based model (EBM) defines a probability density through an energy function $E_\theta: \mathcal{X} \to \mathbb{R}$, typically a neural network with parameters $\theta \in \mathbb{R}^P$, as
\begin{equation}
    p_\theta(\vx) = \frac{\exp(-E_\theta(\vx))}{Z_\theta}, \qquad Z_\theta = \int_{\mathcal{X}} \exp(-E_\theta(\vx)) \,\mathrm{d}\vx,
    \label{eq:ebm}
\end{equation}
where the integral becomes a sum on discrete spaces. EBMs place no structural constraints on $E_\theta$ \citep{lecun2006tutorial}, but the intractable normalising constant $Z_\theta$ makes maximum likelihood estimation difficult. The gradient of the negative log-likelihood takes the contrastive form
\begin{equation}
    \nabla_\theta \, \mathbb{E}_{p_\text{data}}[-\log p_\theta(\vx)] = \mathbb{E}_{p_\text{data}(\vx)}[\nabla_\theta E_\theta(\vx)] - \mathbb{E}_{p_\theta(\vx')}[\nabla_\theta E_\theta(\vx')],
    \label{eq:ml-gradient}
\end{equation}
which requires samples $\vx'$ from the model itself \citep{song2021train}. These are usually approximated with a few steps of Markov chain Monte Carlo (MCMC), in contrastive divergence \citep{hinton2002training} and its persistent \citep{tieleman2008training} and short-run \citep{nijkamp2019learning} variants. For continuous data the standard sampler is stochastic gradient Langevin dynamics (SGLD) \citep{welling2011bayesian}, which needs an input gradient $\nabla_{\vx} E_\theta$ at every step; for discrete data, Gibbs updates or gradient-informed proposals can be used instead \citep{grathwohl2021oops}. Each parameter update therefore contains an inner sampling loop, which dominates the cost of training, and truncated chains give biased gradients that are a well-documented source of instability \citep{du2019implicit, nijkamp2019learning, song2021train}.

\paragraph{Energy discrepancy} Energy discrepancy (ED) \citep{schroder2023energy} is an EBM training objective that needs neither samples from the model nor its score. Given a perturbation kernel $q(\vy \mid \vx)$, define the contrastive potential $E_q(\vy) := -\log \int q(\vy \mid \vx)\exp(-E(\vx))\,\mathrm{d}\vx$. The energy discrepancy between $p_\text{data}$ and an energy $E$ is
\begin{equation}
    \mathrm{ED}_q(p_\text{data}, E) := \mathbb{E}_{p_\text{data}(\vx)}[E(\vx)] - \mathbb{E}_{p_\text{data}(\vx)}\mathbb{E}_{q(\vy \mid \vx)}[E_q(\vy)].
    \label{eq:ed}
\end{equation}
Under mild conditions on $q$, $\mathrm{ED}_q$ is minimised, uniquely up to a constant, by the energy $E^\ast$ with $p_\text{data} \propto \exp(-E^\ast)$ \citep{schroder2023energy}. (\citet{schroder2023energy} denote the energy by $U$; we write $E$ throughout.) For a Gaussian perturbation $\vy = \vx + \sqrt{t}\,\bm{\xi}$, the symmetry of the kernel gives the practical estimator
\begin{equation}
    \mathcal{L}_{t,M,w}(\theta) = \frac{1}{N}\sum_{i=1}^{N} \log\left[\frac{w}{M} + \frac{1}{M}\sum_{j=1}^{M} \exp\!\left(E_\theta(\vx^i) - E_\theta(\vx^i + \sqrt{t}\,\bm{\xi}^i + \sqrt{t}\,\bm{\xi}'^{\,i,j})\right)\right],
    \label{eq:ed-loss}
\end{equation}
with $\bm{\xi}^i, \bm{\xi}'^{\,i,j} \sim \mathcal{N}(0, \mI)$, $M$ negative particles per data point, and a stabilisation offset $w$ that controls the variance of the estimate \citep{schroder2023energy}. Evaluating \eqref{eq:ed-loss} costs $M{+}1$ parallel energy evaluations and a single backward pass per data point, with no sequential sampling loop and no gradients with respect to the inputs, as opposed to contrastive divergence.
The construction carries over to discrete and mixed-type data by replacing the Gaussian with Bernoulli or neighbourhood-based perturbations \citep{schroder2023training}, or with the solution of a heat equation on a graph over each variable's states, which also gives a factorised kernel for mixed-type data \citep{schroder2024heat}. In the form we use numerical features receive Gaussian noise with standard deviation $t_{\text{num}}$, and each categorical feature is kept with probability $e^{-t_{\text{cat}}}$ and otherwise redrawn uniformly from its $C_j$ levels.
\begin{equation}
    q_t(\vy \mid \vx) = \prod_{k=1}^{d_n} \mathcal{N}\!\left(y_k \,\middle|\, x_k,\, t_{\text{num}}^2\right) \; \prod_{j=1}^{d_c} q_{\text{uni}}(y_j \mid x_j), \qquad q_{\text{uni}}(y_j \mid x_j) = e^{-t_{\text{cat}}}\delta(y_j, x_j) + \frac{1-e^{-t_{\text{cat}}}}{C_j},
    \label{eq:mixed-kernel}
\end{equation}
 This is \citet{schroder2024heat}'s uniform heat-equation kernel (TabED-Uni), with the time rescaled so that the keep probability does not depend on $C_j$, as in their implementation. Both factors are symmetric, so one draws a perturbed point $\vy^i \sim q_t(\cdot \mid \vx^i)$ per data point and $M$ recovery candidates $\vx^{i,j}_- \sim q_t(\cdot \mid \vy^i)$, and estimates ED with
\begin{equation}
    \mathcal{L}_{\mathrm{ED}}(\theta) = \frac{1}{N}\sum_{i=1}^{N} \log\left[w + \frac{1}{M}\sum_{j=1}^{M} \exp\!\left(E_\theta(\vx^i) - E_\theta(\vx^{i,j}_-)\right)\right].
    \label{eq:ed-tab}
\end{equation}
We place the offset $w$ next to the particle \emph{mean}, so that the stabilisation does not change with $M$; our $w$ corresponds to $Mw$ in the notation of \citet{schroder2023training, schroder2024heat}, and with all features numerical \eqref{eq:ed-tab} reduces to \eqref{eq:ed-loss} up to $w \mapsto w/M$.

\paragraph{Joint energy-based models}
\label{sec:jem-background}
A standard classifier $f_\theta: \mathcal{X} \to \mathbb{R}^K$ maps an input to logits and defines $p_\theta(y \mid \vx) = \mathrm{softmax}(f_\theta(\vx))[y]$. \citet{grathwohl2019your} observed that the same logits can be read as the negative energy of a joint EBM;
\begin{equation}
    p_\theta(\vx, y) = \frac{\exp(f_\theta(\vx)[y])}{Z_\theta}, \qquad
    E_\theta(\vx) = -\log \sum_{y'} \exp(f_\theta(\vx)[y']),
    \label{eq:jem}
\end{equation}
so that marginalising over $y$ gives a density $p_\theta(\vx) \propto \exp(-E_\theta(\vx))$, while $p_\theta(y \mid \vx)$ remains the usual softmax classifier, unaffected by $Z_\theta$. Training maximises the factorised joint likelihood
\begin{equation}
    \log p_\theta(\vx, y) = \log p_\theta(y \mid \vx) + \log p_\theta(\vx),
    \label{eq:jem-loss}
\end{equation}
where the first term is the cross-entropy objective and the second is an EBM maximum likelihood problem for the marginal energy $E_\theta(\vx)$. The original Joint Energy-based Model (JEM) estimates its gradient with \eqref{eq:ml-gradient} and persistent short-run SGLD ($K \approx 20$ steps per update) with a replay buffer \citep{grathwohl2019your}. Later work improves its stability and cost but keeps the sampling loop \citep{yang2021jempp, yang2023towards} or replaces it with an adversarially trained generator \citep{grathwohl2020no}. Besides generation, joint training has been reported to improve calibration, out-of-distribution detection and adversarial robustness of the classifier \citep{grathwohl2019your, yang2023towards}. JEMs have so far been developed and evaluated almost exclusively on images, so their capabilities as joint models of tabular data are still unknown.

\subsection{Tabular diffusion models}
\label{sec:diffusion-background}
Diffusion models \citep{sohl2015deep, ho2020denoising, song2021score} define a generative model by reversing a fixed noising process. In the continuous case, a data point $\vx_0$ is gradually corrupted into $\vx_\tau = \alpha_\tau \vx_0 + \sigma_\tau \bm{\epsilon}$, $\bm{\epsilon} \sim \mathcal{N}(0, \mI)$, over a time $\tau \in [0, 1]$, until $\vx_1$ is close to pure noise. A neural network $\bm{\epsilon}_\theta(\vx_\tau, \tau)$ is trained to predict the added noise with the denoising objective
\begin{equation}
    \mathcal{L}_{\text{diff}}(\theta) = \mathbb{E}_{\vx_0, \tau, \bm{\epsilon}}\big[\lambda(\tau)\, \lVert \bm{\epsilon}_\theta(\vx_\tau, \tau) - \bm{\epsilon} \rVert^2\big],
    \label{eq:diffusion-loss}
\end{equation}
which, for suitable weights $\lambda(\tau)$, is a variational upper bound on $-\log p_\theta(\vx_0)$ \citep{ho2020denoising, song2021score}. Samples are generated by integrating the learned reverse process from noise, typically with tens to hundreds of network evaluations. 
To be applicable to mixed-type tables, categorical variables need a discrete corruption process, such as multinomial diffusion \citep{hoogeboom2021argmax} or masking. In that case $\bm{\epsilon}$ and $\bm{\epsilon}_\theta$ in \eqref{eq:diffusion-loss} cover only the numerical features: a categorical feature is resampled or masked instead of shifted by noise, and the network outputs a distribution over its categories, whose cross-entropy at the original category replaces the squared error for that feature. 
TabDDPM \citep{kotelnikov2023tabddpm} combines Gaussian diffusion on numerical features with multinomial diffusion on categorical ones. TabSyn \citep{zhang2024tabsyn} first maps each row to a continuous latent vector with a variational autoencoder over the numerical and categorical features and trains a score-based diffusion model \citep{karras2022elucidating} in that latent space. TabDiff \citep{shi2024tabdiff} defines one continuous-time diffusion over the numerical and categorical features, Gaussian for numerical and masking for categorical ones, with learnable, feature-specific noise schedules. 
Both TabSyn and TabDiff are trained for thousands of epochs and are, at the time of writing, among the strongest generators of tabular data \citep{shi2024tabdiff, scassola2026sobering}.

\paragraph{Diffusion models as classifiers} A diffusion model can also be used as a classifier \citep{li2023diffusion}. By Bayes' rule, $p(y \mid \vs) \propto p(\vx \mid y)\,p(y)$ and thus the most probable class for an input $\vx$ is the one that maximises $p(\vx \mid y)\,p(y)$. A diffusion model does not give this likelihood exactly, but its denoising loss \eqref{eq:diffusion-loss} is an upper bound on the negative log-likelihood \citep{kingma2021variational} and can be used in its place. \citet{li2023diffusion} condition an image diffusion model $p_\theta(\vx \mid y)$ on each class in turn, compute the denoising loss of the image for every class with the same noise draws, and predict the class with the lowest loss. Because their image datasets have balanced classes, they take $p(y)$ to be uniform.

\subsection{Tabular foundation models}
\label{sec:tfm-background}
Tabular foundation models classify a new dataset without being trained on it. Prior-data fitted networks (PFNs) \citep{muller2022transformers} are transformers that are pre-trained once on millions of synthetic datasets drawn from a prior over data-generating processes. When faced with a new prediction task the training dataset $\mathcal{D} = \{(\vx_i, y_i)\}_{i=1}^n$ is passed to the network as context together with the test rows, and a single forward pass approximates the posterior predictive distribution $p(y \mid \vx, \mathcal{D})$. This is also called \emph{in-context learning} \citep{brown2020language}: there is no training on $D$ and the cost moves to inference, where every prediction is a forward pass over the whole context, with attention costs that grow quadratically in the number of context rows. 
TabPFN \citep{hollmann2023tabpfn} alternates attention over the rows and over the columns of a table and, in its second version, matches or outperforms tuned gradient-boosted trees on small and medium-sized tables \citep{hollmann2025accurate}. 
TabICL \citep{qu2025tabicl} first embeds each row into a single vector, using attention within each column and then within each row, and only then lets the test rows attend to the labelled training rows. Because this in-context stage operates on one token per row rather than one per cell, TabICL scales to larger tables than TabPFN. TabFM \citep{kong2026tabfm} is a recent model of the same kind that embeds after a number of TabPFN-style alternating attention layers first, making it more expensive but more performant.
All these foundation models are trained as either classifiers and regressors, depending on the type of $y$ in the predictive distribution $p(y \mid \vx, \mathcal{D})$. 

\paragraph{Tabular foundation models as generators} Because a numerical or categorical feature $x$ can itself be treated as the prediction target conditioned on the other features, a regression and a classification model \emph{together} can define every conditional of the table. \citet{hollmann2025accurate} show that this can be used to turn TabPFN into a generative model of whole rows by the chain rule,
\begin{equation}
    p(\vx, y \mid \mathcal{D}) = \prod_{j=1}^{d} q\big(x_j \mid \vx_{<j}, \mathcal{D}_{:, <j}\big)\; q\big(y \mid \vx, \mathcal{D}\big),
    \label{eq:tfm-chain}
\end{equation}
where each factor $q\big(x_j \mid \vx_{<j}, \mathcal{D}_{:, <j}\big)$ is one in-context prediction with the already generated columns as context, and the label is drawn last from the classifier.

\subsection{Probabilistic circuits}
\label{sec:pc-background}
A probabilistic circuit (PC) \citep{choi2020probabilistic} represents a distribution as a computational graph of sum units (weighted mixtures with non-negative weights), product units (factorisations) and input units (simple univariate distributions such as Gaussians or categoricals); sum-product networks are a well-known instance \citep{poon2011sum}. If every sum unit mixes inputs with the same scope (\emph{smoothness}) and every product unit multiplies inputs with disjoint scopes (\emph{decomposability}), the circuit is normalised by construction, and any marginal can be computed exactly in a single forward pass by setting the input units of the marginalised variables to one.
Exact sampling is ancestral, without MCMC. PCs are trained by maximum likelihood, with expectation-maximisation or gradient descent. Their structure can be learned from data, for instance from a Chow--Liu tree \citep{chow1968approximating} over the variables, which gives the hidden Chow--Liu tree (HCLT) \citep{liu2021tractable}, and tensorised implementations scale them to many units per variable \citep{peharz2020einsum, liu2023scaling,loconte2025relationshiptensorfactorizationscircuits}. \citet{scassola2026sobering} recently showed that such a tree-structured PC (TabPC), trained by maximum likelihood, matches or outperforms the diffusion models of \Cref{sec:diffusion-background} on standard tabular generation benchmarks at a fraction of their cost.

\section{Included Methods}
\label{sec:routes}

This section lists every joint tabular modeling method considered in our study, how each one defines its classifier and generator, and how they were made applicable to mixed-type tables. We also list a number of reference baselines for both generation and classification. \Cref{tab:candidates} at the end of the section summarises all candidates and serves as a reference for the rest of the paper.

\subsection{Joint models and choice of candidates}
\label{sec:routes-choice}

\paragraph{Joint models} We call a method \emph{joint} if one fit to the training table yields both a classifier $p(y \mid \vx)$ and a generator of rows $(\vx, y)$. A \emph{pair} fits a classifier and a generator separately (\Cref{sec:routes-two}).
%The candidates differ in how closely their two halves agree: the energy-based models, the PC and the foundation-model generators draw the generated label from the model's own classifier, so that generated labels follow $p(y \mid \vx)$ exactly, while the diffusion models generate $(\vx, y)$ jointly and their classifier inverts the model only approximately.

\paragraph{Four model families} We take our candidates from the four families of \Cref{sec:background}. 
These families differ in how they provide both the classifier and generator.
Probabilistic circuits and joint energy-based models model $p(\vx, y)$ itself. 
Tabular foundation models are classifiers and regressors, and become generators through the chain rule in \eqref{eq:tfm-chain}. 
Tabular diffusion models are generators, and become classifiers through Bayes' rule. 
For each family, \Cref{sec:cand-ebm,sec:cand-diffusion,sec:cand-tfm,sec:cand-pc} describe the candidate methods for this study and what adaptations we apply to yield a joint model of mixed-type tables.

\paragraph{Choice of candidates} We do not attempt to cover every tabular classifier and generator. The study is centred on the recent deep-learning families of \Cref{sec:background}, and per family we chose candidate methods that are: (i) at the time of writing among the strongest published representatives of their family on tabular data, (ii) have public code, and (iii) have a single hyperparameter that controls the bulk of their compute budget, such that we can evaluate compute-performance trade off of each method at different settings. 
The energy-based family has three candidates because they share one backbone and differ only in their training objective, which allows for a controlled comparison of EBM training objectives as a nice side result of this study (\Cref{sec:jem-cost}). 
We exclude GAN- and VAE-based generators such as CTGAN \citep{xu2019modeling}, because these are consistently outperformed by the diffusion models we include \citep{zhang2024tabsyn, shi2024tabdiff}. We also exclude generators based on large language models \citep{borisov2023language}, whose compute lies far outside the budgets that we deem the 'resource constrained setting'. 
%Other joint models of tabular data, such as energy-based boosted trees \citep{bravo2025nrgboost} and further foundation-model generators \citep{margeloiu2024tabebm}, are discussed in \Cref{sec:eval-limitations}.
% Within the energy-based family, three models share one backbone and differ only in their training objective, which gives a controlled comparison of how to train an energy-based joint model. 
% The PC represents the tractable models that were recently shown to be state of the art for tabular generation \citep{scassola2026sobering}. 
% We do not include GAN- and VAE-based generators such as CTGAN \citep{xu2019modeling}, which are consistently outperformed by the diffusion models we include \citep{zhang2024tabsyn, shi2024tabdiff}, nor generators based on large language models \citep{borisov2023language}, whose compute lies far outside the budgets we compare. A missing method is therefore a missing representative of its family, and the protocol of \Cref{sec:experiments} can be extended with further candidates without changing the comparison.

% \paragraph{How candidates are described} In the following sections, we state for every candidate joint modeling method how it classifies, how it generates, how it handles mixed features, what we adapted, and its computational budget hyperparameter. Hyperparameter configurations and budget grids are listed in \Cref{app:budgets}.

\subsubsection{Energy-based models}
\label{sec:cand-ebm}
\label{sec:routes-direct}

We include three EBM-based methods that use a similar architecture: a feed-forward network $f_\theta$ with a learned embedding per categorical column, so that it is defined on mixed-type inputs $\vx = (\vx^{\text{num}}, \vx^{\text{cat}})$. Samples are drawn from these EBMs during training and for downstream generation using the same Langevin-within-Gibbs scheme: SGLD updates for the numerical features and exact Gibbs resampling of each categorical feature from its conditional at the cost of one forward pass \emph{per category}. This combines the Langevin-within-Gibbs sampler of \citet{kelly2021directly}, where the discrete variable is the label, with the Langevin and Gibbs updates of the contrastive-divergence baseline of \citet{schroder2024heat}. 
All MCMC chains are initialized from the same prior: independent standard normal numerical features and uniformly drawn category levels. Below we describe the three EBM candidates, which differ in their training objective.

\paragraph{JEM-SGLD \citep{grathwohl2019your}} The original JEM, which estimates the gradient of $\log p_\theta(\vx)$ in \eqref{eq:jem-loss} with persistent SGLD and a replay buffer. To make it applicable to mixed-type data, we used the Langevin-Gibbs sampler described above during training and downstream inference. The main budget hyperparameter is the number of MCMC steps per parameter update $K$ where each step costs one forward pass and one input gradient for the numerical features plus one forward pass per category level ($\sum_j C_j$ in total) for the Gibbs sweep, so with categorical features the cost per step grows with the total number of levels.

\paragraph{JEM-ED} A JEM in which the maximum-likelihood term for the marginal energy $E_\theta(\vx)$ in \eqref{eq:jem-loss} is replaced by energy discrepancy, with the mixed perturbation kernel of \citet{schroder2024heat} (\eqref{eq:mixed-kernel}). The classifier is trained with cross-entropy, so the objective is
\begin{equation}
    \mathcal{L}(\theta) = \mathcal{L}_{\mathrm{ED}}(\theta) + \alpha \, \mathcal{L}_{\mathrm{clf}}(\theta),
    \label{eq:jem-ed-objective}
\end{equation}
with $\mathcal{L}_{\mathrm{ED}}$ being the estimator from \eqref{eq:ed-tab}, $\mathcal{L}_{\mathrm{clf}}$ the cross-entropy of $p_\theta(y \mid \vx)$, and $\alpha$ the weight of the discriminative term. As the budget hyperparameter, we use the number of negative particles $M$. One update evaluates the energy on the clean batch and on $M$ perturbed model copies in parallel, followed by one backward pass so it needs no input gradients, no replay buffer and no sampler hyperparameters. MCMC is only required during downstream generation tasks, and thus its cost is decoupled from training.

\paragraph{TabED-joint \citep{schroder2024heat}} The purely generative parameterisation of \citet{schroder2024heat} that uses one energy $E_\theta(\vx, y)$ over the joint table, with the label embedded as just another categorical column, trained with \eqref{eq:ed-tab} on the whole table (the label included) and without a cross-entropy term. Its architecture thus has only a single output unit. 
Classification is the softmax of $-E_\theta(\vx, c)$ over the classes $c$, and generation runs MCMC on the marginal energy $-\log\sum_c \exp(-E_\theta(\vx, c))$ and labels the samples with this classifier. 
%Both need one forward pass per class wherever the energy of a row is evaluated, against one for a JEM. Note that this makes that TabED-joint requires $C\,\times$ the amount of forward passes during generation over JEM-SGLD and JEM-ED. 
Its main budget hyperparameter is the number of particles $M$, as for JEM-ED.

\subsubsection{Tabular diffusion models}
\label{sec:cand-diffusion}
\label{sec:routes-gen2clf}

Tabular diffusion models are developed as generators of entire table rows $(\vx, y)$ and require a classifier $p(y \mid \vx)$ to be considered a joint model. We obtain one by transferring the diffusion classifier (\citep{li2023diffusion}, see \Cref{sec:diffusion-background}) to tabular diffusion models of mixed-type data. For a test row $\vx$ and every candidate class $c$, we form the row $(\vx, y{=}c)$ and estimate its per-row denoising loss
\begin{equation}
    L_c(\vx) = \frac{1}{S}\sum_{s=1}^{S} \big\lVert \bm{\epsilon}_\theta(\vz_{\tau_s}, \tau_s) - \bm{\epsilon}_s \big\rVert^2, \qquad \vz_0 = (\vx, y{=}c),
    \label{eq:diff-clf-loss}
\end{equation}
over $S = 250$ noise draws $(\tau_s, \bm{\epsilon}_s)$ shared by all classes, so that the comparison between classes is not dominated by Monte Carlo noise. For $L_c$ we use a uniform weighting, instead of the weighting $\lambda(\tau)$ used during training. The class probabilities are $p(y{=}c \mid \vx) \propto \exp(-L_c(\vx)/T)$, with the temperature $T$ fitted on the validation split. When $L_c$ bounds $-\log p(\vx \mid c)$ rather than $-\log p(\vx, c)$, we multiply by the class prior, because tabular labels are often imbalanced. Every classification thus costs a set of denoising passes per class, whereas generating a row requires just one.

To our knowledge, tabular diffusion models have not been evaluated as classifiers before. We apply this construction to the two strongest published tabular diffusion models at the time of writing. For both, the main budget hyperparameter is the number of training epochs.

\paragraph{TabDiff \citep{shi2024tabdiff}} TabDiff adds Gaussian noise to the numerical columns and noises the categorical columns, $y$ included, by masking. Masked categorical columns have no noise-prediction error; in $L_c$ they contribute the cross-entropy of recovering their masked values, weighted as in the likelihood bound of masked diffusion \citep{sahoo2024simple, shi2024simplified}. During training the label is masked like any other categorical column, but in the classifier we keep it unmasked, so that it acts as the condition, as in \citet{li2023diffusion}. Its $L_c$ then bounds $-\log p(\vx \mid c)$, and we add the class prior.

\paragraph{TabSyn \citep{zhang2024tabsyn}} TabSyn encodes each row, the label included, into a continuous latent vector with a variational autoencoder and trains the diffusion model in that latent space. In \eqref{eq:diff-clf-loss}, $\vz_0$ is therefore the VAE encoding of $(\vx, y{=}c)$. Because the label is part of the encoded row, its $L_c$ bounds the joint $-\log p(\vx, c)$ and already contains the prior, as opposed to TabDiff.

\subsubsection{Tabular foundation models}
\label{sec:cand-tfm}
\label{sec:routes-clf2gen}

Tabular foundation models are in-context classifiers and regressors (\Cref{sec:tfm-background}), which we turn into generators using the chain rule from \eqref{eq:tfm-chain}. For every column $x_j$ in turn, the rows being generated are passed as queries to the classifier (categorical columns) or the regressor (numerical columns), with the context restricted to the columns generated so far, and the column's value is sampled from the predictive distribution $q\big(x_j \mid \vx_{<j}, \mathcal{D}_{:, <j}\big)$. The label $y$ is generated last by the same classifier that is used for prediction, so generated labels follow the model's own $p(y \mid \vx, \mathcal{D})$. 
We construct the classifier and the regressor as two separately pre-trained networks instead of two output heads of one backbone, following \citet{hollmann2025accurate}. Note that we still count them as one joint model, since conditioning both networks on the same context table is a single fit that produces both a generator and a classifier.
In terms of computational cost, fitting is free since the foundation model is pre-trained and assumed to be publicly available. Classification costs one in-context forward pass, and generation one forward pass per column, each batched over query rows (\Cref{app:cost}). Categorical features are ordinal-encoded as inputs and drawn from the classifier's distribution over their categories during generation in the chain rule setup from \eqref{eq:tfm-chain}.

\textbf{TabPFN} \citep{hollmann2025accurate} The appendix of \citet{hollmann2025accurate} proposes the generative construction in \eqref{eq:tfm-chain} but provides no evaluation beyond an illustration.
We implement it ourselves, with three adaptations: we use one column order instead of an average over orders, we implement the first column $x_1$ in generation as a draw from its empirical distribution in the context instead of predicted from a dummy noise column, and we use quantile-normalised numerical features. In preliminary experiments our implementation was at least as good as the official {implementation}\footnote{\url{https://github.com/PriorLabs/tabpfn-extensions}}, which we observed to mishandle low-cardinality numerical columns. To our knowledge, this is the first quantitative evaluation of TabPFN as a generator. The budget hyperparameter is the size of the context. 

\textbf{TabICL} \citep{qu2025tabicl} We implement the chain-rule based generator for TabICL using the official {code}\footnote{\url{https://github.com/soda-inria/tabicl/blob/main/src/tabicl/_unsupervised/unsupervised.py}}, which we use as is, with the label appended as the last column and the context capped at 2{,}000 rows. The classifier uses the full training split as context. The budget hyperparameter is the number of ensemble members, for both the generator and the classifier.

\paragraph{Energy-based alternative (ablation)} A classifier with logits $f(\vx)$ also defines a class-conditional energy $E(\vx \mid y) = -f(\vx)[y]$, as in a JEM, and TabPFGen \citep{ma2024tabpfgen} samples TabPFN this way with SGLD, without explicitly maximizing likelihood of $\vx$. We evaluate TabPFGen and the same construction on a differentiable re-implementation of TabICL as an ablation study in \Cref{app:sampler-ablation}. We observed that when initialized from a prior, the sampling chains did not converge to high quality $\vx$, and thus the chain rule based generators are better on every generative metric at a fraction of the cost.

\subsubsection{Probabilistic circuits}
\label{sec:cand-pc}
A PC is a tractable model of $p(\vx, y)$ (see \Cref{sec:pc-background}), so it is a joint model without any adaptation: the label is just another categorical variable of the circuit and the circuit is fitted by maximum likelihood on the joint table. Numerical features enter through Gaussian input units, and categorical features through categorical input units. Classification computes the exact conditional $p(y{=}c \mid \vx) = p(\vx, c) / \sum_{c'} p(\vx, c')$, at the cost of one forward pass per class. Generation draws $(\vx, y)$ by exact ancestral sampling, which costs one categorical draw per variable and row and needs neither MCMC nor a diffusion process.

We follow \textbf{TabPC} \citep{scassola2026sobering}: a Chow-Liu tree is learned over all variables of the joint table, rooted at its centre and compiled into a smooth and decomposable circuit in our own JAX implementation, which is trained with TabPC's optimisation recipe. We depart from TabPC in two implementation choices: (i) the we use the hidden Chow-Liu tree layout of \citet{liu2021tractable} instead of TabPC's CP layers, and (ii) we use the preprocessing that is shared by all candidates in this study (\Cref{sec:experimental-setup}) instead of that in the original paper. The budget hyperparameter is the number of units per variable, $K_{\text{PC}}$.

\subsection{Single-task references and the two-model baseline}
\label{sec:routes-two}

\paragraph{Single-task references} We add a number of sole classification and generation baselines to the study to add context to the computation-performance trade-offs of the joint models.
We include a simple Multilayer Perceptron (\textbf{MLP}) as reference classifier, which is the JEM backbone trained with the cross-entropy term alone.
We also include \textbf{XGBoost} \citep{chen2016xgboost} with native categorical splits and early stopping. Finally, we add \textbf{TabFM} \citep{kong2026tabfm}: an in-context classifier with ordinal-encoded categorical features as the state-of-the-art for tabular classification at the time of this project.
For generation, we add two references that bound the generation score rather than compete for it, following the guidelines in \citet{scassola2026sobering}: a \textbf{fully factorised sampler (FF)}, which draws every column, the label included, independently from its training marginal, and \textbf{real data}, a sample of training rows of the size of the test split. FF reproduces every marginal but no dependency between columns, and real data is the best any generator can score on our metrics; together they provide the reasonable lower and upper bound for generative metrics.

\paragraph{The two-model baseline.} A joint model is only worth it if it is at least as good as the obvious alternative: a classifier and a generator fitted separately. We build this baseline by pairing every classifier with every generator in our study (except a method with itself, also at another budget). The computational cost of a pair is defined as the training and prediction cost of its classifier plus the training and sampling cost of its generator. So for example, we evaluate the combined cost and performance of an MLP for classification and TabDiff at a certain hyperparameter setting for generation. For more details see \Cref{sec:experimental-setup}.

\subsection{Summary}
\label{sec:routes-summary}

\Cref{tab:candidates} summarises all candidates. Keep in mind that results that rely on one of our adaptations of a published method are technically novel methods, so weak results may reflect the adaptation rather than the original published method. We included a summary of these adaptations for each method in the table. We also added a summarized explanation of how categorical features are handled in the method. 'Native purpose' states the functionality the method provides in its original publication. Configurations and budget grids can be found in \Cref{app:budgets}.

\begin{table}[t]
  \centering
  \caption{All candidates of the study, grouped by model family.}
  \label{tab:candidates}
  \scriptsize
  \setlength{\tabcolsep}{3pt}
  \renewcommand{\arraystretch}{1.15}
  \begin{tabularx}{\textwidth}{@{}>{\raggedright\arraybackslash}p{1.55cm} >{\raggedright\arraybackslash}p{0.95cm} >{\raggedright\arraybackslash}X >{\raggedright\arraybackslash}X >{\raggedright\arraybackslash}X >{\raggedright\arraybackslash}X >{\raggedright\arraybackslash}p{1.45cm}@{}}
    \toprule
    Method & Native purpose & Classifier & Generator & Categorical features & What is new or adapted & Budget hyperparameter \\
    \midrule
    \multicolumn{7}{@{}l}{\emph{Energy-based models}} \\
    JEM-ED & joint & softmax of logits & MCMC on $E_\theta(\vx)$, $y \sim p_\theta(y|\vx)$ & embeddings; uniform ED kernel & ED for the JEM marginal objective & particles $M$ \\
    JEM-SGLD & joint & softmax of logits & as JEM-ED & embeddings; Gibbs in every SGLD step & Gibbs sampling in training & SGLD steps $K$ \\
    TabED-joint & joint & softmax of $-E_\theta(\vx, c)$ over classes $c$ (one pass per class) & as JEM-ED & embeddings; uniform ED kernel (label too) & - & particles $M$ \\
    \midrule
    \multicolumn{7}{@{}l}{\emph{Tabular diffusion models}} \\
    TabDiff & generator & diffusion classifier & reverse diffusion & masking & diffusion classifier for tables & epochs \\
    TabSyn & generator & diffusion classifier (latent) & latent reverse diffusion $+$ decoder & VAE encoding & as TabDiff & epochs \\
    \midrule
    \multicolumn{7}{@{}l}{\emph{Tabular foundation models}} \\
    TabPFN & classifier & in context & chain rule & ordinal encoding & ordinal encoding; own implementation of the published generator & context size \\
    TabICL & classifier & in context & chain rule & as TabPFN & ordinal encoding & ensemble size \\
    \midrule
    \multicolumn{7}{@{}l}{\emph{Probabilistic circuits}} \\
    PC (HCLT) & joint & exact conditional & exact ancestral sampling & categorical input units & HCLT layout; shared preprocessing & units $K_{\text{PC}}$ \\
    \midrule
    \multicolumn{7}{@{}l}{\emph{Single-task references}} \\
    MLP & classifier & softmax of logits & -- & embeddings & -- & -- \\
    XGBoost & classifier & boosted trees & -- & native splits & -- & rounds \\
    TabFM & classifier & in context & -- & ordinal & ordinal encoding & ensemble size \\
    FF & generator & -- & from training marginal & -- & -- & -- \\
    Real data & generator & -- & sample of training rows & -- & -- & -- \\
    \bottomrule
  \end{tabularx}
\end{table}

\section{Experiments}
\label{sec:experiments}
Our goal is to determine which of the joint modeling methods described in the previous section gives the best classifier and generator in one model, at what cost, and whether a single model is ever preferable over two separate ones at a certain computational budget. The experiments are organised around four questions:
\begin{itemize}
  \item \textbf{Efficiency (\Cref{sec:tradeoff})} Which joint model provides the best prediction, generation and overall performance at different computational budgets?
  \item \textbf{Strengths per metric (\Cref{sec:main-results})} On which predictive and generative metrics is each joint model strong or weak, at its best budget?
  \item \textbf{One model or two (\Cref{sec:one-vs-two})} At a certain compute budget, is a single joint model ever better than a classifier and a generator trained separately?
  \item \textbf{Numerical vs mixed type data (\Cref{sec:num-vs-mixed})} How does the performance of each joint model change when the data is purely numerical versus mixed-type?
\end{itemize}

\subsection{Experimental setup}
\label{sec:experimental-setup}

\paragraph{Benchmarks}
We use two benchmark suites (details in \Cref{app:datasets}).
\emph{UCI-mixed} contains five UCI classification datasets: \textit{Adult, Default, Diabetes} and \textit{Online Shoppers} have numerical and categorical features, and \textit{MAGIC} is purely numerical.
The \emph{OpenML-CC18} benchmark \citep{bischl2021openml} is used through 24 classification datasets with at most 2{,}000 rows, which is $6\times$ less than the smallest dataset in UCI-mixed. We additionally split the suite into an 18-dataset purely numerical subset and a 6-dataset mixed-type subset (\Cref{app:datasets}).

Note that UCI-mixed can be seen as the larger table benchmark, while CC18 is the small table benchmark in this study. 
Every method is run with five seeds; each seed re-draws a stratified 64/16/20 train/validation/test split.

\paragraph{Preprocessing} All methods share the same preprocessing of the data. Numerical features are mapped to an approximately standard normal scale through a Gaussian quantile transformation fitted on the training split and categorical features are represented by integer codes. How each method then treats the categorical codes is described per candidate in \Cref{sec:routes}.

\paragraph{Hyperparameters}
Each candidate is swept over a grid of its 'budget hyperparameter' (see \Cref{tab:candidates}). All other hyperparameters are fixed across datasets at the authors' published or package defaults for the published methods and at one shared setting for the energy-based models. \Cref{tab:main} reports each method at its best budget (\Cref{sec:main-results}). The budget grids, the default budgets, all other hyperparameter settings and implementation details can be found in \Cref{app:hyperparameters}.

\paragraph{Metrics}
Predictive performance is measured on the held-out test split with accuracy, one-vs-rest AUROC, the multiclass Brier score \citep{brier1950verification}, the conditional negative log-likelihood (NLL) and the expected calibration error (ECE \citealp{guo2017calibration}).
For generative performance metrics we follow \citet{scassola2026sobering}. They show that the metrics most commonly used to rank tabular generators, TREND (pairwise correlation similarity) and C2ST with a logistic-regression classifier, are near-saturated for state-of-the-art diffusion models and equally close to perfect for the FF baseline that samples every column independently from the marginal. SHAPE, the third common metric, only compares univariate marginals. As replacements they propose a weighted normalised mutual-information similarity (wNMIS), which measures pairwise dependence between generated variables of any type, and C2ST with an XGBoost classifier, which detects discrepancies in the full joint distribution. 
Therefore, we let each method generate a synthetic table of both $\vx$ and $y$ of the size of the test split and report wNMIS, C2ST-XGB and also the machine-learning efficacy (MLE), which is the test AUROC of an XGBoost classifier trained on the synthetic table and evaluated on real data \citep{xu2019modeling}. 
For a joint model MLE is interesting because it is the only one of the three metrics that measures whether the generated \emph{labels} actually depend on the generated features. A sampler that produced perfect rows with random labels would score close to the ceiling on wNMIS and C2ST-XGB but at chance on MLE. We do not use SHAPE, TREND or C2ST-LR because of the objections raised by \citep{scassola2026sobering}.

\paragraph{Aggregate scores}
In order to decide whether one joint method is better than another on either prediction or generation, a single score per task is needed.
We therefore formulate a single aggregate prediction and aggregate generation score. 
The metrics that evaluate either prediction or generation listed above cannot simply be averaged, since they differ in domain, scale and trivial upper and lower bounds. For example, a naive majority-class classifier reaches an accuracy of $0.73$ on UCI-mixed and a generator with independent columns obtains a wNMIS of $0.80$ but a C2ST-XGB of $0.02$ (\Cref{tab:references}), so that C2ST-XGB would decide an unnormalised average. Therefore, we rescale every metric between a trivial \emph{floor} (score $0$) and an ideal \emph{ceiling} (score $1$). For prediction, the floor is the class prior and the ceiling a perfect classifier; for generation, the floor is a fully factorised sampler and the ceiling a sample of real training rows (as suggested by \citet{scassola2026sobering}). This gives two aggregate scores:
\begin{itemize}
  \item \textbf{Prediction score (Pred.):} the mean of $(\text{acc} - \text{acc}_0)/(1 - \text{acc}_0)$, $(\text{AUROC} - 0.5)/0.5$ and $1 - \text{Brier}/\text{Brier}_0$, where subscript 0 denotes the \emph{class prior}, which predicts the empirical class frequencies for every row; a perfect classifier (accuracy and AUROC $1$, Brier score $0$) scores $1$ on each term. The Brier score covers calibration because unlike the NLL, it is bounded so no single over-confident dataset dominates and unlike ECE, it is a proper scoring rule.
  \item \textbf{Generation score (Gen.):} the mean of $(m - m_{\text{FF}})/(m_{\text{real}} - m_{\text{FF}})$ over $m \in \{\text{wNMIS}, \text{C2ST-XGB}, \text{MLE}\}$, where the floor 'FF' is a fully factorised sampler that draws every column independently from its empirical marginal distribution \citep{scassola2026sobering}, and the ceiling is a sample of real training rows of the size of the test split.
\end{itemize}
We compute both scores separately for every dataset and seed, using the floor and ceiling of that same split. We then average them over the five seeds and take the interquartile mean (IQM; \citealp{agarwal2021deep}) of the resulting per-dataset scores over the datasets of a suite.
Additionally we obtain $95\%$ intervals using a bootstrap over datasets (\Cref{tab:scores-ci}) and, for every single model against a pair, the probability $P$ that the pair has the higher overall score, defined below (\Cref{sec:one-vs-two}).
Note that the metrics for the references (class prior, FF sampler, real data) are given in \Cref{app:references}
Our conclusions about which joint modeling method is perceived to perform better on either generation or prediction than another method is based on these two scores. We also add an \emph{overall score}, the geometric mean of the prediction and generation scores, to rank joint modeling methods overall.

\paragraph{Cost accounting}
We define computational cost of a method as the total compute needed to fit it on a new dataset, predict the test split labels and generate a synthetic table the size of the test split.
We report this cost in FLOPs, which are independent of hardware, and in wall-clock time such that we also capture memory-bound and sequential work (e.g the difference between JEM-SGLD and JEM-ED). 
We count all floating-point arithmetic of fitting, prediction and sampling: matrix products are two FLOPs per multiply-accumulate and all other operations are one FLOP per element. We count them for every run with our own counter, which records every operation executed in JAX and PyTorch. XGBoost runs on CPU, so its cost comes from a model fitted to the floating-point instructions it executes. \Cref{app:cost} gives the details.
Wall-clock times are measured on one NVIDIA GH200 per run. Because we recognize the errors that come with counting flops, our claims rest only on cost differences of at least about $2\times$.

\subsection{Efficiency: performance against computational cost}
\label{sec:tradeoff}

\begin{figure}[t]
  \centering
  \includegraphics[width=\textwidth]{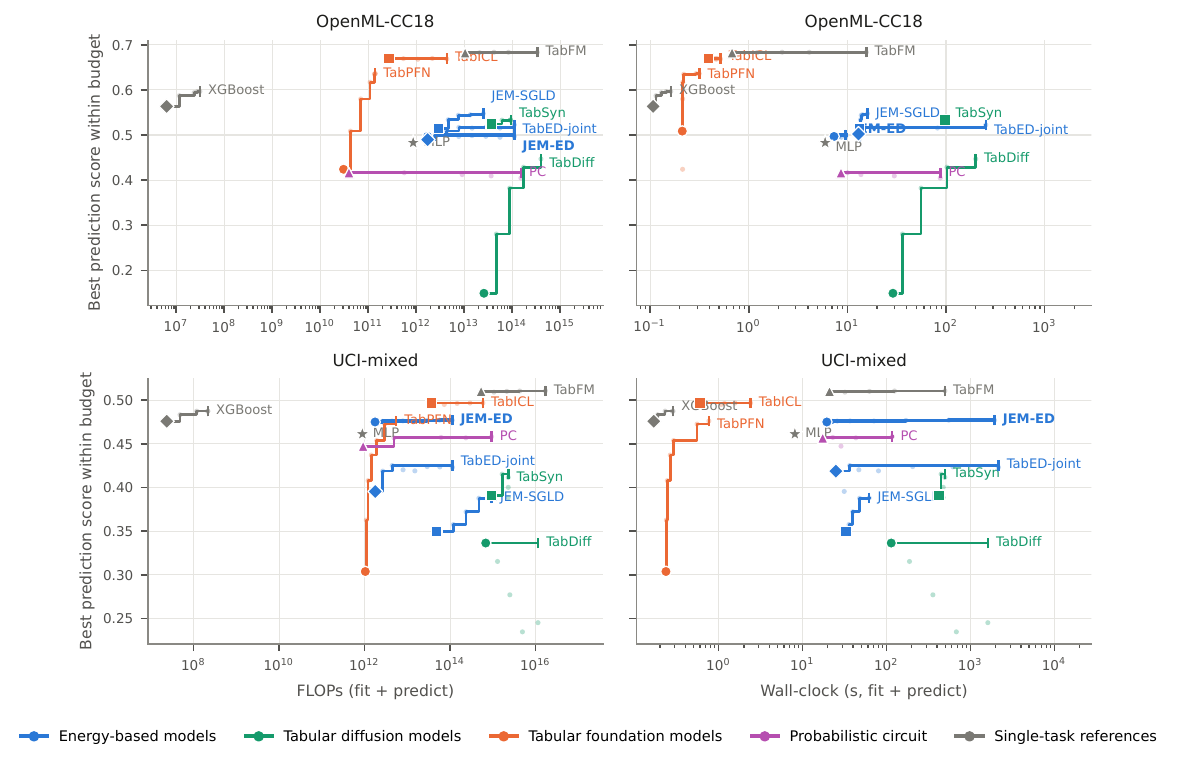}
  \caption{Prediction score (y-axis) against cost of fitting and predicting (x-axis) for every method on OpenML-CC18 (top) and UCI-mixed (bottom). Left: cost as FLOPs of fitting and prediction; right: cost measured as wall-clock time. Each curve represents the scores a method achieves at different settings of its budget hyperparameter. The upper envelope of all curves is the Pareto front.}
  \label{fig:tradeoff-pred}
\end{figure}

\begin{figure}[t]
  \centering
  \includegraphics[width=\textwidth]{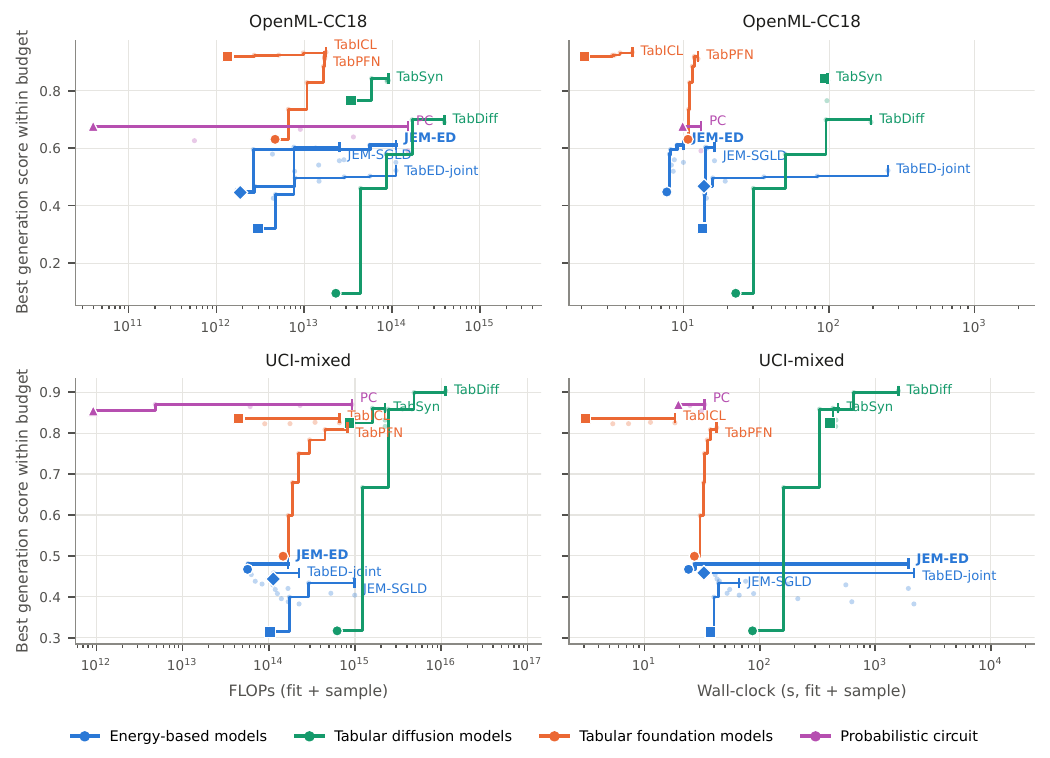}
  \caption{Generation score (y-axis) against the cost of fitting and sampling (x-axis) with left cost as FLOPs and right with cost as wall-clock).}
  \label{fig:tradeoff-gen}
\end{figure}

\begin{figure}[t]
  \centering
  \includegraphics[width=\textwidth]{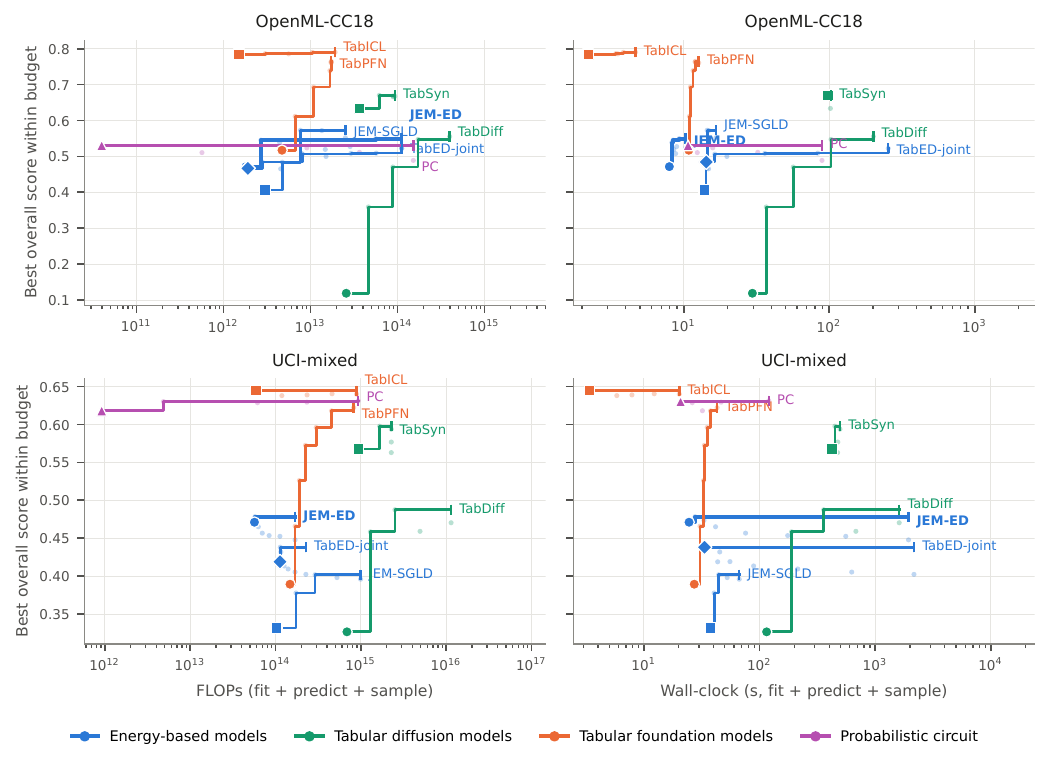}
  \caption{Overall score (geometric mean of the prediction and generation scores) on the y-axis against the total cost of fitting, prediction and sampling (x-axis).}
  \label{fig:tradeoff-overall}
\end{figure}

\Cref{fig:tradeoff-pred,fig:tradeoff-gen,fig:tradeoff-overall} show the aggregated scores that we defined in the previous section for each method, on each benchmark suite, against its cost in FLOPs and wall-clock time, at different settings of its budget hyperparameter given in \cref{tab:main}. 
% Each score is charged only the phases it needs: the prediction score the cost of fitting and prediction, the generation score that of fitting and sampling, and the overall score the total of all three. A joint model's position on the prediction panels therefore does not include its sampler, which accounts for up to $99\%$ of the wall-clock time of the foundation models.

\paragraph{Prediction} 
Looking at the results in \Cref{fig:tradeoff-pred}, we see a similar picture on both the CC18 and UCI-mixed benchmark suites. The Pareto front is formed by XGBoost at the lowest budgets, which reaches its best prediction score of $0.60$ on CC18 at $3.1\cdot10^{7}$ FLOPs ($0.49$ on UCI-mixed at $1.2\cdot10^{8}$), followed by the tabular foundation models at higher computational budgets. On CC18 these are TabPFN ($0.64$ at $1.3\cdot10^{11}$ FLOPs), TabICL with one ensemble member ($0.67$ at $2.7\cdot10^{11}$) and TabFM ($0.68$ from $1.0\cdot10^{13}$ FLOPs on). On UCI-mixed these are TabICL ($0.50$ at $3.6\cdot10^{13}$) and TabFM ($0.51$ from $5.3\cdot10^{14}$). 
As for wall-clock time, XGBoost is the fastest strong classifier, below $0.3$\,seconds on both suites. On CC18, TabICL and TabFM offer a little more performance than XGBoost from respectively $0.4$ and $0.7$\,seconds on, and on UCI-mixed these are preferable at $0.6$ and $21$ seconds onwards, respectively.

\paragraph{Generation} From \Cref{fig:tradeoff-gen} we can conclude that on CC18, the generation Pareto front consists of only two methods: the PC at $4.0\cdot10^{10}$ FLOPs (score $0.68$), and TabICL from a computational budget of $1.3\cdot10^{12}$ FLOPs on (score $0.92$--$0.93$), which no other method reaches at any budget. Notably, tabular diffusion methods cannot touch the generation performance of tabular foundation models on the relatively small datasets of CC18. 
On UCI-mixed the Pareto front is the PC up to $9.2\cdot10^{14}$ FLOPs ($0.85$ at 64 units, $0.87$ from 256 units on), after which TabDiff catches up: at 2{,}000 epochs ($2.4\cdot10^{15}$ FLOPs) it reaches $0.86$, and at 4{,}000 epochs ($4.8\cdot10^{15}$ FLOPs, about $5\times$ the largest PC) it overtakes the PC with a score of $0.90$. 
In terms of wall-clock time, TabICL is the clear winner on CC18, at $2$ to $5$ seconds. On UCI-mixed the generation Pareto front in seconds is TabICL ($0.84$ at $3.1$ seconds), the PC ($0.87$ at $20$ seconds) and TabDiff ($0.90$ at $650$ seconds).
In general it is noticeworthy that tabular foundation models are better at generation than tabular diffusion models after being fit on small tables.
%In terms of wall-clock time, TabICL exceeds a score of $0.8$ at a faster generation time than the PC. 

\paragraph{Overall} The overall score (\Cref{fig:tradeoff-overall}) combines both generation and prediction and is the main 'joint modeling' performance indicator. 
On CC18 its Pareto front consists of the smallest PC ($0.53$ at $4.0\cdot10^{10}$ FLOPs) and from $1.5\cdot10^{12}$ FLOPs on TabICL leads with a score of $0.78$. The next best are TabPFN ($0.76$) and TabSyn ($0.67$). On UCI-mixed the front is the PC at 64 and 256 units ($0.62$ and $0.63$ at $0.9\cdot10^{12}$ and $4.9\cdot10^{12}$ FLOPs), followed by TabICL with one ensemble member ($0.64$ at $5.9\cdot10^{13}$ FLOPs). TabPFN reaches $0.62$ at its largest context of 4{,}096 rows, TabSyn reaches $0.60$, and the energy-based models stay below $0.48$ at any budget.
In terms of wall-clock time, TabICL is the sole winner on CC18, being both the most performant and the fastest joint model: with one ensemble member it reaches $0.78$ in $2.2$\,seconds, against for example $12.7$\,seconds for TabPFN's $0.76$ and $10.7$\,seconds for the smallest PC. 
On UCI-mixed TabICL is again the sole winner, with $0.64$ in $3.4$\,seconds. The PC, which precedes it on the FLOPs front, needs $21$\,seconds for $0.63$, because its fit takes many small sequential updates that FLOPs do not reflect, and TabPFN needs $43$\,seconds for a similar score of $0.62$.

%We can see that the conclusion on which joint modeling provides the best performance-efficiency trade off depends on dataset size: on the small tables of CC18 the foundation models are the most efficient for both prediction and generation, while on larger tables of UCI-7 the PC is the most efficient generator and a diffusion model buys the last three points of generation score ($0.90$ vs.\ $0.87$) at about $10^3\times$ the cost of a PC with 256 units.
Looking at overall competence as a joint model, we conclude that in this study, the PC is the best choice on smaller compute budgets and TabICL at larger budgets, with the gap between the two shrinking for larger training datasets. TabICL is the best choice overall when wall-clock time is most important.

\subsection{Strengths per metric}
\label{sec:main-results}
The previous section compares the methods on the aggregate scores, but in this section we will zoom in a bit more on the individual metrics to evaluate on which aspects of prediction and generation each joint model is strong or weak. \Cref{tab:main} shows these results. We will ignore computational budget in this section. Each method is shown at its \emph{best budget}: the budget with the highest overall score among the budget hyperparameter values we swept (prediction score for the prediction-only methods; ties go to the cheaper budget), regardless of cost.

\begin{table}[t]
  \centering
  \caption{Quality at each method's best budget (highest overall score, or prediction score for prediction-only methods, among the swept budgets), grouped by model family. 
  Every entry is the interquartile mean over the suite's datasets of the per-dataset mean over 5 seeds. 
  Pred.\ and Gen.: prediction and generation scores, normalised per dataset so that the reference models (grey rows) score $0$ (class prior; fully factorised sampler) and $1$ (real data). 
  Overall: their geometric mean; $95\%$ intervals in \Cref{tab:scores-ci}. 
  TFLOP: total FLOPs (fit $+$ predict $+$ sample, $10^{12}$) at that budget.
  Best per suite and column in bold.}
  \label{tab:main}
  \scriptsize
  \setlength{\tabcolsep}{2.6pt}
  \resizebox{\textwidth}{!}{\begin{tabular}{ll|cccccc|cccc|c|r}
\toprule
& & \multicolumn{6}{c|}{Prediction} & \multicolumn{4}{c|}{Generation} & & \\
Method & Budget & Acc $\uparrow$ & AUROC $\uparrow$ & Brier $\downarrow$ & NLL $\downarrow$ & ECE $\downarrow$ & Pred. $\uparrow$ & wNMIS $\uparrow$ & C2ST-XGB $\uparrow$ & MLE $\uparrow$ & Gen. $\uparrow$ & Overall $\uparrow$ & TFLOP \\
\midrule
\multicolumn{14}{l}{\textbf{UCI-mixed (5 classification datasets)}} \\
\multicolumn{14}{l}{\quad\textit{Reference rows (define 0 and 1 of the scores)}} \\
\textcolor{gray}{Class prior (floor)} & -- & \textcolor{gray}{0.729} & \textcolor{gray}{0.500} & \textcolor{gray}{0.388} & \textcolor{gray}{0.576} & -- & \textcolor{gray}{0.000} & -- & -- & -- & -- & -- & -- \\
\textcolor{gray}{Fully factorised (floor)} & -- & -- & -- & -- & -- & -- & -- & \textcolor{gray}{0.804} & \textcolor{gray}{0.019} & \textcolor{gray}{0.505} & \textcolor{gray}{0.000} & -- & -- \\
\textcolor{gray}{Real data (ceiling)} & -- & -- & -- & -- & -- & -- & -- & \textcolor{gray}{0.994} & \textcolor{gray}{0.996} & \textcolor{gray}{0.865} & \textcolor{gray}{1.000} & -- & -- \\
\multicolumn{14}{l}{\quad\textit{Energy-based models}} \\
JEM-ED & 2 & 0.854 & 0.872 & 0.213 & 0.342 & 0.015 & 0.476 & 0.918 & 0.160 & 0.811 & 0.481 & 0.478 & 57.4 \\
JEM-SGLD & 10 & 0.827 & 0.836 & 0.254 & 0.481 & 0.042 & 0.372 & 0.914 & 0.104 & 0.811 & 0.434 & 0.402 & 290 \\
TabED-joint & 2 & 0.833 & 0.851 & 0.240 & 0.380 & 0.024 & 0.419 & 0.921 & 0.156 & 0.797 & 0.459 & 0.438 & 113 \\
\multicolumn{14}{l}{\quad\textit{Tabular diffusion models}} \\
TabDiff & 2000 & 0.813 & 0.788 & 0.264 & 0.410 & 0.023 & 0.277 & 0.975 & 0.734 & \textbf{0.855} & 0.858 & 0.488 & 2{,}502 \\
TabSyn & 1250 & 0.844 & 0.842 & 0.230 & 0.381 & 0.021 & 0.415 & \textbf{0.983} & 0.725 & 0.852 & 0.861 & 0.598 & 1{,}657 \\
\multicolumn{14}{l}{\quad\textit{Tabular foundation models}} \\
TabPFN & 4096 & 0.853 & 0.869 & 0.215 & 0.344 & 0.012 & 0.476 & \textbf{0.983} & 0.591 & 0.847 & 0.813 & 0.622 & 822 \\
TabICL & 1 & 0.862 & 0.880 & 0.202 & 0.325 & 0.013 & 0.497 & 0.982 & 0.652 & 0.844 & 0.836 & \textbf{0.644} & 59.2 \\
\multicolumn{14}{l}{\quad\textit{Probabilistic circuits}} \\
PC (HCLT) & 4096 & 0.847 & 0.854 & 0.225 & 0.360 & 0.013 & 0.458 & 0.981 & \textbf{0.793} & 0.838 & \textbf{0.870} & 0.631 & 930 \\
\multicolumn{14}{l}{\quad\textit{Single-task references}} \\
MLP (CE only) & -- & 0.853 & 0.868 & 0.215 & 0.350 & 0.019 & 0.461 & -- & -- & -- & -- & -- & 0.887 \\
XGBoost & 25 & 0.854 & 0.875 & 0.213 & 0.341 & \textbf{0.010} & 0.487 & -- & -- & -- & -- & -- & $1.2{\times}10^{-4}$ \\
TabFM & 32 & \textbf{0.867} & \textbf{0.886} & \textbf{0.192} & \textbf{0.307} & 0.014 & \textbf{0.511} & -- & -- & -- & -- & -- & $1.7{\times}10^{4}$ \\
\midrule
\multicolumn{14}{l}{\textbf{OpenML-CC18 (24 datasets)}} \\
\multicolumn{14}{l}{\quad\textit{Reference rows (define 0 and 1 of the scores)}} \\
\textcolor{gray}{Class prior (floor)} & -- & \textcolor{gray}{0.572} & \textcolor{gray}{0.500} & \textcolor{gray}{0.522} & \textcolor{gray}{0.828} & -- & \textcolor{gray}{0.000} & -- & -- & -- & -- & -- & -- \\
\textcolor{gray}{Fully factorised (floor)} & -- & -- & -- & -- & -- & -- & -- & \textcolor{gray}{0.891} & \textcolor{gray}{0.226} & \textcolor{gray}{0.520} & \textcolor{gray}{0.000} & -- & -- \\
\textcolor{gray}{Real data (ceiling)} & -- & -- & -- & -- & -- & -- & -- & \textcolor{gray}{0.976} & \textcolor{gray}{0.971} & \textcolor{gray}{0.865} & \textcolor{gray}{1.000} & -- & -- \\
\multicolumn{14}{l}{\quad\textit{Energy-based models}} \\
JEM-ED & 64 & 0.829 & 0.894 & 0.277 & 0.858 & 0.121 & 0.495 & 0.925 & 0.587 & 0.821 & 0.610 & 0.550 & 56 \\
JEM-SGLD & 10 & 0.832 & 0.893 & 0.245 & 0.493 & 0.077 & 0.544 & 0.939 & 0.483 & 0.801 & 0.603 & 0.573 & 7.64 \\
TabED-joint & 128 & 0.820 & 0.891 & 0.257 & 0.504 & 0.068 & 0.522 & 0.924 & 0.573 & 0.740 & 0.522 & 0.522 & 112 \\
\multicolumn{14}{l}{\quad\textit{Tabular diffusion models}} \\
TabDiff & 8000 & 0.797 & 0.855 & 0.286 & 0.479 & 0.060 & 0.447 & 0.946 & 0.690 & 0.810 & 0.690 & 0.555 & 396 \\
TabSyn & 1250 & 0.836 & 0.894 & 0.246 & 0.437 & 0.058 & 0.533 & 0.965 & 0.774 & \textbf{0.840} & 0.843 & 0.670 & 61.6 \\
\multicolumn{14}{l}{\quad\textit{Tabular foundation models}} \\
TabPFN & 4096 & 0.854 & 0.926 & 0.196 & 0.325 & 0.047 & 0.636 & \textbf{0.970} & 0.868 & 0.839 & 0.920 & 0.765 & 17.3 \\
TabICL & 16 & 0.874 & 0.937 & 0.174 & 0.290 & 0.044 & 0.668 & 0.968 & \textbf{0.883} & 0.833 & \textbf{0.934} & \textbf{0.790} & 19.1 \\
\multicolumn{14}{l}{\quad\textit{Probabilistic circuits}} \\
PC (HCLT) & 64 & 0.784 & 0.839 & 0.294 & 0.489 & 0.078 & 0.416 & 0.949 & 0.743 & 0.753 & 0.676 & 0.531 & 0.0397 \\
\multicolumn{14}{l}{\quad\textit{Single-task references}} \\
MLP (CE only) & -- & 0.822 & 0.894 & 0.308 & 1.523 & 0.146 & 0.483 & -- & -- & -- & -- & -- & 0.865 \\
XGBoost & 500 & 0.840 & 0.911 & 0.225 & 0.389 & 0.056 & 0.597 & -- & -- & -- & -- & -- & $3.1{\times}10^{-5}$ \\
TabFM & 4 & \textbf{0.881} & \textbf{0.941} & \textbf{0.164} & \textbf{0.275} & \textbf{0.042} & \textbf{0.684} & -- & -- & -- & -- & -- & 41.7 \\
\bottomrule
\end{tabular}}
\end{table}

\paragraph{Energy-based models}
On the larger UCI-mixed tables, JEM-ED is the best EBM method on every prediction metric. All three are especially weak generators in terms of C2ST-XGB: their samples are easy to tell apart from real data (C2ST-XGB at most $0.16$, against $0.996$ for real data) when compared to the diffusion models and foundation models. Hence their generation scores of $0.43$-$0.48$ come mostly from a good wNMIS and MLE score, with again JEM-ED outperforming the others.
On the smaller CC18 tables the picture changes in two ways. First, all three generate more realistic rows with C2ST-XGB between $0.48$ and $0.59$, with JEM-ED again having the most realistic ($0.587$) though still far off of foundation and diffusion models. 
Second, the ranking on prediction reverses: JEM-SGLD is the best calibrated of the three (Brier score $0.245$ and NLL $0.49$, against $0.277$ and $0.86$ for JEM-ED) and therefore has the highest prediction score ($0.54$ against $0.50$ for JEM-ED and $0.52$ for TabED-joint). Noticeworthy is that TabED-joint has the weakest label dependence on CC18 (MLE $0.740$, $64\%$ of the way from chance to real data, against $81$--$87\%$ for the JEMs), because its labels are sampled from the same joint energy as the features rather than from a separate classifier trained with cross-entropy.

Against the reference classifiers, the energy-based models fare well on the large tables and less so on the small ones. On UCI-mixed, JEM-ED obtains almost identical metrics as XGBoost and overall on prediction it performs slightly better than the MLP ($0.476$ vs. $0.461$). On the smaller tables of CC18 all three are clearly behind XGBoost, mostly because of calibration (NLL $0.49$--$0.86$ against $0.39$; ECE $0.07$--$0.12$ against $0.06$).

The MLP baseline has the same backbone and number of updates as the EBMs but is trained with cross-entropy only, so comparing it with them gives an indication on what modeling $\vx$ jointly does to the classifier. On UCI-mixed the effect is small (JEM-ED: accuracy $+0.1$ points, NLL $0.342$ against MLP's $0.350$). 
On CC18 the generative term improves calibration of the EBMs significantly compared to the MLP. 
The generative term therefore seems acts as a regulariser of the classifier rather than as a tax on it on smaller datasets, consistent with \citet{schroder2024heat,grathwohl2019your}. Note that little to no preventative measures were implemented for overfitting (e.g early stopping, dropout, weight decay) which may influence the results.

\paragraph{Probabilistic circuit}
What the PC is good at depends strongly on the size of the tables. On UCI-mixed it is the best overall generator of all joint models at their best budgets and a competent classifier: its accuracy ($0.847$) is close to the MLP's ($0.853$) and XGBoost's ($0.854$), and its calibration is among the best (ECE $0.013$, against $0.010$ for XGBoost and $0.012$ for TabPFN).
%Its weak point as a generator is label dependence: although its rows are the hardest to tell apart from real data compared to the other methods, 
On the small CC18 tables the PC is the weakest classifier among the joint models with an aggregate prediction score of $0.42$, against at least $0.45$ for every other joint model and only a mediocre generator with a generation score of $0.68$, against at least $0.84$ for TabSyn, TabPFN and TabICL.
We observed that the budget behaves differently too: on CC18 the smallest circuit (64 units) is the best, and larger ones generate worse (generation score $0.63$ at 256 units), whereas on UCI-mixed the largest circuit (4{,}096 units) is the best. So its performance is very dependent on having the right capacity for the dataset, which should be evaluated with BIC for example.

\paragraph{Tabular foundation models}
On the small CC18 tables, whose training splits fit entirely in context of both TabPFN and TabICL, both models are strong at both prediction and generation. TabICL is ahead if TabPFN on every prediction metric and on the aggregate generation score and both tabular foundation models notably have a C2ST-XGB of $0.883$ and $0.868$ against $0.774$ for TabSyn, the best generator outside this family, so the generated data that is significantly harder to distinguish from real data. The generated labels are highly correlated with the generated features (MLE $91$--$92\%$ of the way from chance to real data), because the label is drawn last from the model's own classifier; only TabSyn's are slightly better ($93\%$).

On the larger UCI-mixed tables, TabICL remains the strongest classifier among the joint models (accuracy $0.862$, aggregate prediction score $0.50$), while TabPFN is on par with JEM-ED (prediction score $0.48$ for both). As generators, the tabular foundation models start to fall behind: showing a C2ST-XGB $0.652$ for TabICL and $0.591$ for TabPFN, against $0.725$--$0.793$ for the PC and the diffusion models, and generation scores $0.84$ and $0.81$ against $0.86$ to $0.87$. Their label dependence stays high, so foundation models seem to mostly lose the edge in realism of generated features on large tables, not the link between features and label. An obvious cause may be context: the UCI training sets have $8{,}000$ to $65{,}000$ rows, far more than the context size of the foundation models. However we observe that TabPFN improves with every doubling of its context up to 2{,}048 rows, but then levels off: its overall score is $0.596$, $0.618$ and $0.622$ at 1{,}024, 2{,}048 and 4{,}096 rows, still below the $0.644$ that TabICL reaches with one ensemble member at about a fourteenth of the FLOPs.

\paragraph{Tabular diffusion models}
On UCI-mixed the diffusion models are, together with the PC, the best generators, and notably their generated labels are more correlated with the generated features than those of any other joint model (MLE $96$--$97\%$ of the way from chance to real data).
As classifiers, however, they are the weakest joint models on these larger tables: TabSyn is on par with TabED-joint, and TabDiff has the lowest aggregate prediction score of all ($0.28$).
On CC18, TabSyn remains a strong generator, second only to the foundation models (generation score $0.84$), and its generated labels are again the most feature-dependent of all joint models. 
TabDiff falls back to the level of the PC as a generator ($0.69$). 
As classifiers both stay behind: TabSyn is close to the energy-based models, and TabDiff is ahead of only the PC. Their classifiers do have a distinctive profile. On CC18 they are better calibrated than the energy-based models, the PC and the MLP, with an ECE of about $0.06$ that is close to XGBoost's: they make more errors than the strongest classifiers, but are rarely confidently wrong, as one may expect from class probabilities derived from a likelihood bound rather than from a loss driven to zero on the training data. 
Ranking is their weak point: on UCI-mixed TabDiff has the lowest AUROC of all joint models ($0.788$), below even its accuracy ($0.813$), and TabSyn's AUROC is also slightly below its accuracy, whereas for every other model the AUROC is above the accuracy.

How performance reacts to longer training depends on the size of the tables. 
On CC18, training TabDiff longer improves both generation and prediction: from 500 to 8{,}000 epochs its prediction score rises from $0.15$ to $0.45$ and its generation score from $0.10$ to $0.69$. 
On UCI-mixed however the performance on the tasks diverges: the generation score keeps rising up to 8{,}000 epochs (from $0.32$ to $0.90$), while the prediction score falls from the start (from $0.34$ to $0.25$). TabDiff's best overall budget (2{,}000 epochs) is therefore neither its best classifier nor its best generator. TabSyn shows neither effect, because its early stopping ends the training of its latent diffusion model before about 2{,}500 epochs, so beyond its best budget of 1{,}250 epochs its scores change little and without a trend.
So once the tables are large enough, a diffusion model is a joint model whose two halves want different amounts of training: the longer it fits the data distribution, the less its ELBO separates the classes, and a single budget cannot serve both tasks. 
Classification is also the expensive half of these models: reading out a class takes an ELBO per class, estimated from 250 diffusion-time draws per row, which costs six to fourteen times the FLOPs of generating the synthetic table.

\subsection{One model or two?}
\label{sec:one-vs-two}

\begin{figure}[t]
  \centering
  \includegraphics[width=\textwidth]{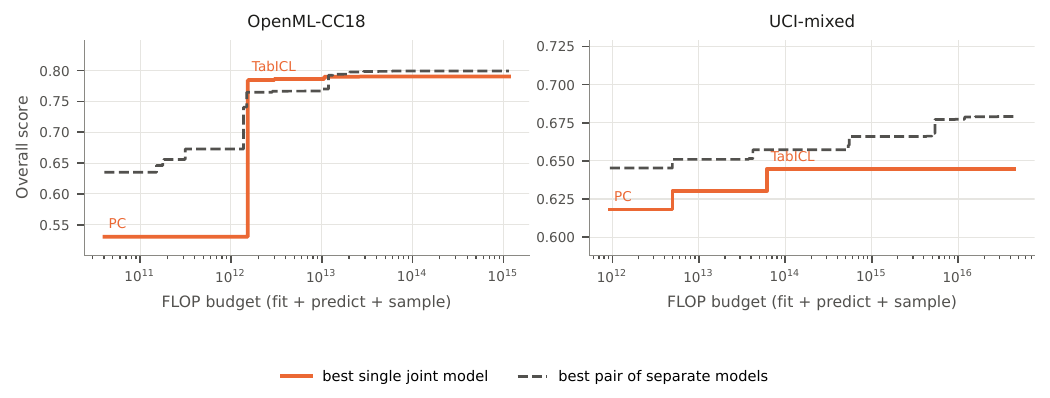}
  \caption{The best single joint model (solid line) and the best pair of a separately trained classifier and generator (dashed line) that fit within a given FLOP budget (x-axis), both chosen by their overall score (y-axis).}
  \label{fig:ovt-simple}
\end{figure}

\begin{table}[t]
  \centering
  \caption{Per FLOP budget, the best single joint model and the best pair of separate models within the budget, both chosen by overall score. $P$: bootstrap probability over datasets that the pair has the higher overall score. A pair may reuse one half of a joint model. Models are stated in 'method\,@\,budget' format (\Cref{app:budgets}).}
  \label{tab:ovt-simple}
  \scriptsize
  \resizebox{\textwidth}{!}{\begin{tabular}{ll|lccc|lccc|c}
\toprule
& & \multicolumn{4}{c|}{Best single joint model} & \multicolumn{4}{c|}{Best pair of separate models} & \\
Suite & FLOPs & model & Pred. & Gen. & Overall & models & Pred. & Gen. & Overall & $P$ \\
\midrule
OpenML-CC18 & $\leq 10^{11}$ & PC\,@64 & 0.416 & 0.676 & 0.531 & XGBoost\,@500 + PC\,@64 & 0.597 & 0.676 & 0.635 & 1.00 \\
 & $\leq 10^{12}$ & PC\,@64 & 0.416 & 0.676 & 0.531 & TabICL\,@2 + PC\,@64 & 0.669 & 0.676 & 0.673 & 1.00 \\
 & $\leq 10^{13}$ & TabICL\,@2 & 0.669 & 0.924 & 0.786 & TabPFN\,@2048 + TabICL\,@8 & 0.636 & 0.932 & 0.770 & 0.01 \\
 & $\leq 10^{14}$ & TabICL\,@16 & 0.668 & 0.934 & 0.790 & TabFM\,@4 + TabICL\,@16 & 0.684 & 0.934 & 0.799 & 1.00 \\
 & any & TabICL\,@16 & 0.668 & 0.934 & 0.790 & TabFM\,@4 + TabICL\,@16 & 0.684 & 0.934 & 0.799 & 1.00 \\
\midrule
UCI-mixed & $\leq 10^{12}$ & PC\,@64 & 0.447 & 0.854 & 0.618 & XGBoost\,@25 + PC\,@64 & 0.487 & 0.854 & 0.645 & 1.00 \\
 & $\leq 10^{13}$ & PC\,@256 & 0.457 & 0.870 & 0.630 & XGBoost\,@25 + PC\,@256 & 0.487 & 0.870 & 0.651 & 1.00 \\
 & $\leq 10^{14}$ & TabICL\,@1 & 0.497 & 0.836 & 0.644 & TabICL\,@1 + PC\,@256 & 0.497 & 0.870 & 0.657 & 0.77 \\
 & any & TabICL\,@1 & 0.497 & 0.836 & 0.644 & TabFM\,@32 + TabDiff\,@8000 & 0.511 & 0.903 & 0.679 & 1.00 \\
\bottomrule
\end{tabular}}
\end{table}

In this section we ask whether, at a given computational budget, one joint model is worth using instead of a separate classifier and generator. A pair has to share the computational budget between two models, whereas a joint model can spend all of it on one more powerful model. For each FLOP budget, \Cref{tab:ovt-simple} lists the best single joint model and the best pair of separate models that fit within it, both chosen by their overall score, and \Cref{fig:ovt-simple} shows the same comparison over all budgets. A pair may combine any classifier with any generator of another method. It may therefore also reuse one half of a joint model, for example TabICL's generator next to a separate classifier, as a practitioner would.
We compare the two on the overall score, which weighs prediction and generation score equally. Because the differences are often small, we use the bootstrap probability $P$ over datasets that the pair has the higher overall score: the pair is better if $P \geq 0.95$, the joint model if $P \leq 0.05$, and the two are tied in between.

\paragraph{Results}
On CC18 (upper half of \Cref{tab:ovt-simple}, left panel of \Cref{fig:ovt-simple}), a joint model is better than the best pair only at medium budgets. Below $1.5\cdot10^{12}$ FLOPs the smallest PC is the best joint model, but XGBoost next to the same PC is better ($0.64$ against $0.53$; $P = 1.00$). From $1.5\cdot10^{12}$ FLOPs on, TabICL is the best joint model (overall score $0.78$--$0.79$). Up to $1.2\cdot10^{13}$ FLOPs it is better than or tied with every pair ($P \leq 0.07$), because no classifier stronger than its own fits within the budget. With more compute, a TabFM classifier next to TabICL's generator first ties with TabICL and is better from $2.8\cdot10^{13}$ FLOPs on ($0.80$ against $0.79$; $P = 1.00$). The gain is small but consistent across datasets, because both use the same generator.

On UCI-mixed a pair is at least as good as the best joint model at every budget. Below $5.9\cdot10^{13}$ FLOPs, XGBoost next to the PC is better than the PC alone ($0.65$ against $0.62$--$0.63$; $P = 1.00$). From $5.9\cdot10^{13}$ FLOPs on, TabICL with one ensemble member is the best joint model ($0.64$). It is tied (statistically) with the pair of its own classifier and the PC ($0.66$; $P = 0.77$). From $4.9\cdot10^{15}$ FLOPs on, a pair with TabDiff as the generator is better ($0.67$--$0.68$; $P = 1.00$), mostly through generation ($0.90$ against $0.84$).

The best pairs explain the result. Each joint model is weak on one half, and the cheapest way to strengthen that half is to 'buy' it from a dedicated model. A classifier like XGBoost costs far less than a generator, so the best pair often keeps the generator of a joint model and adds a separate classifier. On large tables the PC is a cheap, strong generator, but a poorer classifier than the in-context models. On small tables TabICL is strong on both halves, so at medium budgets it is the best option on both as a joint model. 
We note that a single joint model has further benefits that our scores do not capture, such as one set of weights to maintain and a classifier that is consistent with the generator by construction, as it is for the column-by-column generators and the JEMs.

\subsection{Numerical versus mixed-type data}
\label{sec:num-vs-mixed}
In this section we compare performance of all methods on numerical and mixed-type datasets separately, to see whether the inclusion of categorical type features matters for the relative performance of the methods.
\begin{table}[t]
  \centering
  \caption{Performance at best budget hyperparameter on the CC18 numerical (18 datasets) and mixed-type (6 datasets) subsets, given as in \Cref{tab:main}. Best per subset and column in bold.}
  \label{tab:split}
  \scriptsize
  \setlength{\tabcolsep}{3pt}
  \resizebox{\textwidth}{!}{\begin{tabular}{l|rccccccc|rccccccc}
\toprule
& \multicolumn{8}{c|}{CC18 numerical (18)} & \multicolumn{8}{c}{CC18 mixed (6)} \\
Method & Budget & Acc & Brier & Pred. & C2ST & MLE & Gen. & Overall & Budget & Acc & Brier & Pred. & C2ST & MLE & Gen. & Overall \\
\midrule
JEM-ED & 2 & 0.848 & 0.263 & 0.567 & 0.394 & 0.848 & 0.484 & 0.524 & 4 & 0.726 & 0.416 & 0.321 & 0.848 & \textbf{0.732} & 0.853 & 0.523 \\
JEM-SGLD & 10 & 0.846 & 0.229 & 0.590 & 0.326 & 0.837 & 0.554 & 0.572 & 10 & 0.724 & 0.355 & 0.379 & 0.820 & 0.680 & 0.746 & 0.532 \\
TabED-joint & 128 & 0.838 & 0.238 & 0.572 & 0.419 & 0.769 & 0.433 & 0.498 & 8 & 0.731 & 0.356 & 0.385 & 0.871 & 0.689 & 0.874 & 0.580 \\
TabDiff & 8000 & 0.815 & 0.263 & 0.494 & 0.534 & 0.839 & 0.608 & 0.548 & 8000 & 0.716 & 0.361 & 0.326 & 0.902 & 0.713 & 0.932 & 0.551 \\
TabSyn & 1250 & 0.855 & 0.221 & 0.594 & 0.731 & 0.870 & 0.853 & 0.712 & 10000 & 0.720 & 0.365 & 0.350 & 0.899 & 0.702 & 0.827 & 0.538 \\
TabPFN & 4096 & 0.878 & 0.164 & 0.707 & 0.849 & 0.879 & 0.926 & 0.809 & 2048 & 0.719 & 0.356 & 0.373 & \textbf{0.934} & 0.682 & 0.917 & 0.585 \\
TabICL & 8 & 0.892 & 0.152 & 0.722 & \textbf{0.863} & \textbf{0.884} & \textbf{0.927} & \textbf{0.818} & 16 & \textbf{0.752} & 0.312 & 0.448 & 0.919 & 0.699 & \textbf{0.957} & \textbf{0.655} \\
PC (HCLT) & 64 & 0.821 & 0.253 & 0.509 & 0.680 & 0.807 & 0.698 & 0.596 & 64 & 0.657 & 0.434 & 0.192 & 0.846 & 0.604 & 0.599 & 0.339 \\
MLP (CE only) & -- & 0.844 & 0.284 & 0.562 & -- & -- & -- & -- & -- & 0.707 & 0.528 & 0.222 & -- & -- & -- & -- \\
XGBoost & 500 & 0.858 & 0.205 & 0.650 & -- & -- & -- & -- & 500 & 0.737 & 0.329 & 0.404 & -- & -- & -- & -- \\
TabFM & 4 & \textbf{0.896} & \textbf{0.142} & \textbf{0.738} & -- & -- & -- & -- & 8 & \textbf{0.752} & \textbf{0.311} & \textbf{0.450} & -- & -- & -- & -- \\
\bottomrule
\end{tabular}}
\end{table}

\Cref{tab:split} splits CC18 into its 18 purely numerical and 6 mixed-type datasets. Note that for each method we chose the budget that produced the best performance. One thing that stands out generally is that prediction scores are higher on numerical data than on mixed-type data, while for generation the opposite holds.
It is also interesting that EBMs are the weakest generators on numerical data but competitive on mixed-type data, where TabED-joint ($0.87$) and JEM-ED ($0.85$) even obtain higher generation scores than TabSyn ($0.83$). 
TabDiff and TabSyn swap places: TabSyn is the better generator on numerical data ($0.85$ against $0.61$ for generative score) while TabDiff, which diffuses categorical columns directly, is better on mixed-type data ($0.93$ against $0.83$). 
As classifiers, all methods are much closer on mixed-type data, where almost all reach an accuracy of $0.72$--$0.75$. TabPFN, second only to TabICL on numerical data, falls back to the level of the EBMs and diffusion models there, possibly because it receives categorical features as integer codes. 
Energy discrepancy brings no classification advantage over JEM-SGLD on mixed-type data, since our JEM-SGLD adaptation now also handles categorical features natively through Gibbs updates. However one must keep in mind that computational efficiency is a significant advantage of JEM-ED ($5$--$32\times$ fewer training FLOPs, see \Cref{tab:jem-cost}), because of the cost of Gibbs sampling when categorical data becomes involved. With only six mixed-type datasets, these differences are only indicative.

\subsection{Limitations of the evaluation}
\label{sec:eval-limitations}

We point out the current limitations of the empirical study.
\textbf{Scope:} The empirical study is not complete in terms of methods that are compared. Many more methods from literature or different architectures for the methods present (e.g the specific type of PC used) could be added. However, we believe we chose a representative subset of methods from 4 different families of joint modeling methods that are close to state-of-the-art on generation or prediction. Moreover, some of the joint modeling methods we present here have not been proposed or studied before, which is why we regard this study a first step towards a more complete empirical evaluation of joint modeling methods instead of a complete benchmark.
\textbf{Aggregation:} Scores are interquartile means over datasets with bootstrap intervals over datasets. With five datasets in UCI-mixed and six in CC18-mixed these intervals are wide, and several 'joint-vs-pair' verdicts have bootstrap probabilities between $0.4$ and $0.7$, so differences of a few hundredths of a score should not be over-interpreted. Normalising per dataset is also unstable when the real data barely beats the floor on a metric (for instance wNMIS on tic-tac-toe, where real data and the fully factorised sampler differ by $0.006$), which can push a dataset's score far outside $[0, 1]$.
\textbf{Scores:} label dependence is measured only through MLE, which reflects whether a classifier trained on the synthetic data transfers to real data, not directly whether the generated labels follow the real conditional distribution. A class-conditional two-sample test would measure that more directly.
\textbf{Hyperparameters:} no method was tuned per dataset (\Cref{sec:experimental-setup}) but mostly uses default hyperparameters. This favours methods whose defaults were tuned on similar data. 
\textbf{Tasks not covered:} out-of-distribution detection and adversarial robustness, which motivate JEMs (see \citep{grathwohl2019your}), are not part of this study and the joint models were only evaluated on classification datasets, not regression.
% \textbf{Baseline budgets:} TabPFN's context is swept up to 4{,}096 rows, below the training-set size on UCI-mixed, where its scores barely rise beyond 2{,}048 rows (overall score $0.618$ to $0.622$); TabICL's generator sees at most 2{,}000 rows, and its budget hyperparameter, the number of ensemble members, changes its cost but not its fidelity. TabICL's generator in particular may therefore be under-powered on UCI-mixed, and a larger context may change the answer there.
\textbf{Cost:} The way we measure FLOPs is not exact due to computational constraints and because it is nontrivial to validate the FLOP counters in general and relatively between methods. FLOPs are only a proxy for compute (\Cref{app:cost}). They miss memory-bound work, and they do not show the effect of parallelization. For example, every SGLD step in JEM-SGLD is counted, but these steps run sequentially. JEM-ED may use the same amount of counted flops on its negative particles, but run this compute in parallel, which can speed up execution.
\section{Conclusion}
\label{sec:conclusions}

We asked what the best and most efficient way is to obtain one model that both classifies and generates tabular data, and whether one model is ever worth it compared with two specialists. We compared joint models from four families: energy-based models, probabilistic circuits, tabular foundation models and tabular diffusion models. 
Some of these joint modeling methods were based on existing adjustments and techniques, though have never been introduced or evaluated before this work.
We evaluated all joint modeling methods' performance on 29 classification datasets along a budget hyperparameter that we chose, scoring prediction and generation per dataset between a trivial floor and real data.

\paragraph{Which joint model at which budget}
On smaller training datasets from CC18, the smallest probabilistic circuit is the best joint model on lower FLOP budgets and from there on TabICL, a tabular foundation model that generates column by column with its own predictive distributions, is the best based on overall score.
On the larger UCI-mixed tables, we see the same pattern. 
Looking at prediction and generation separately, TabICL is the best classifier among the joint models on both small and large tables, while XGBoost is the cheapest strong classifier overall. For generation the answer depends on the size of the training dataset: the foundation models are the best generators on small tables, whereas on large tables the probabilistic circuit is the most efficient generator and a diffusion model the best one when compute is plentiful.

\paragraph{Joint models versus pairs}
A joint model is rarely preferable to two separate models. It beats every pair only on small tables at medium budgets, where TabICL is both the best generator and classifier. In all other cases a pair showed itself to be at least as good. The best pair often keeps the generator of a joint model and adds a cheap separate classifier such as XGBoost, which improves prediction at little extra cost. A single model remains preferable mainly where its other benefits matter: one set of weights and a classifier that is consistent with its generator.

\paragraph{How the families differ}
The tabular foundation models are the strongest joint classifiers and the strongest generators on small training tables. On large tables their limited context holds them back as generators. The diffusion models are among the best generators on large tables but weak classifiers, and there their classifier and generator need different amounts of training. 
The probabilistic circuit is a cheap and strong generator, provided its capacity matches the size of the data. 
The energy-based models are competitive classifiers whose generative term seems to act as a regulariser, but are weak generators of numerical features in terms of especially C2ST-XGB. Within this family, energy discrepancy reaches the quality of SGLD training at up to two orders of magnitude fewer training FLOPs.

\section*{Author contributions}
Tim d'Hondt designed the study, implemented experiments and wrote the paper. Oleksii Furman designed the study, implemented experiments. Jan Poręba and Damian Kwolek implemented experiments. Mykola Pechenizkiy and Maciej Zieba supervised the project.

\section*{Acknowledgments}
Oleksii Furman and Maciej Zi\k{e}ba's work was supported by the National Science Centre (Poland) Grant No. 2024/55/B/ST6/02100.

This work was partly supported by H2020 SmartChange project, funded within EU’s Horizon Europe research program (GA No. 101080965) and EMDAIR exploratory project VACE.

We gratefully acknowledge Polish high-performance computing infrastructure PLGrid (HPC Center: ACK Cyfronet AGH) for providing computer facilities and support within computational grant no. PLG/2026/019381

\bibliographystyle{plainnat}
\bibliography{references}

@inproceedings{yang2023towards,
  title={Towards bridging the performance gaps of joint energy-based models},
  author={Yang, Xiulong and Su, Qing and Ji, Shihao},
  booktitle={Proceedings of the IEEE/CVF Conference on Computer Vision and Pattern Recognition},
  pages={15732--15741},
  year={2023}
}

@article{grathwohl2019your,
  title={Your classifier is secretly an energy based model and you should treat it like one},
  author={Grathwohl, Will and Wang, Kuan-Chieh and Jacobsen, J{\"o}rn-Henrik and Duvenaud, David and Norouzi, Mohammad and Swersky, Kevin},
  journal={arXiv preprint arXiv:1912.03263},
  year={2019}
}

@article{schroder2023energy,
  title={Energy discrepancies: a score-independent loss for energy-based models},
  author={Schr{\"o}der, Tobias and Ou, Zijing and Lim, Jen and Li, Yingzhen and Vollmer, Sebastian and Duncan, Andrew},
  journal={Advances in Neural Information Processing Systems},
  volume={36},
  pages={45300--45338},
  year={2023}
}

@article{schroder2023training,
  title={Training Discrete Energy-Based Models with Energy Discrepancy},
  author={Schr{\"o}der, Tobias and Ou, Zijing and Li, Yingzhen and Duncan, Andrew B},
  journal={arXiv preprint arXiv:2307.07595},
  year={2023}
}

@article{qu2025tabicl,
  title={{TabICL}: A tabular foundation model for in-context learning on large data},
  author={Qu, Jingang and Holzm{\"u}ller, David and Varoquaux, Ga{\"e}l and Le Morvan, Marine},
  journal={arXiv preprint arXiv:2502.05564},
  year={2025}
}

@article{kong2026tabfm,
  title={{TabFM}: A zero-shot foundation model for tabular data},
  author={Kong, Weihao and Ilan, Erez Louidor and Nie, Shuxin and Narayan, Taman and Sen, Rajat and Zhou, Yichen and Fu, Deqing and Oymak, Samet and Das, Abhimanyu},
  journal={arXiv preprint arXiv:2609.37959},
  year={2026}
}

@article{hinton2002training,
  title={Training products of experts by minimizing contrastive divergence},
  author={Hinton, Geoffrey E},
  journal={Neural Computation},
  volume={14},
  number={8},
  pages={1771--1800},
  year={2002}
}

@inproceedings{tieleman2008training,
  title={Training restricted {B}oltzmann machines using approximations to the likelihood gradient},
  author={Tieleman, Tijmen},
  booktitle={Proceedings of the 25th International Conference on Machine Learning},
  pages={1064--1071},
  year={2008}
}

@inproceedings{nijkamp2019learning,
  title={Learning non-convergent non-persistent short-run {MCMC} toward energy-based model},
  author={Nijkamp, Erik and Hill, Mitch and Zhu, Song-Chun and Wu, Ying Nian},
  booktitle={Advances in Neural Information Processing Systems},
  volume={32},
  year={2019}
}

@inproceedings{nijkamp2020anatomy,
  title={On the anatomy of {MCMC}-based maximum likelihood learning of energy-based models},
  author={Nijkamp, Erik and Hill, Mitch and Han, Tian and Zhu, Song-Chun and Wu, Ying Nian},
  booktitle={Proceedings of the AAAI Conference on Artificial Intelligence},
  volume={34},
  number={04},
  pages={5272--5280},
  year={2020}
}

@inproceedings{welling2011bayesian,
  title={Bayesian learning via stochastic gradient {L}angevin dynamics},
  author={Welling, Max and Teh, Yee Whye},
  booktitle={Proceedings of the 28th International Conference on Machine Learning},
  pages={681--688},
  year={2011}
}

@inproceedings{yang2021jempp,
  title={{JEM}++: Improved techniques for training {JEM}},
  author={Yang, Xiulong and Ji, Shihao},
  booktitle={Proceedings of the IEEE/CVF International Conference on Computer Vision},
  pages={6494--6503},
  year={2021}
}

@misc{loconte2025relationshiptensorfactorizationscircuits,
      title={What is the Relationship between Tensor Factorizations and Circuits (and How Can We Exploit it)?}, 
      author={Lorenzo Loconte and Antonio Mari and Gennaro Gala and Robert Peharz and Cassio de Campos and Erik Quaeghebeur and Gennaro Vessio and Antonio Vergari},
      year={2025},
      eprint={2409.07953},
      archivePrefix={arXiv},
      primaryClass={cs.LG},
      url={https://arxiv.org/abs/2409.07953}, 
}

@inproceedings{grathwohl2021oops,
  title={Oops {I} took a gradient: Scalable sampling for discrete distributions},
  author={Grathwohl, Will and Swersky, Kevin and Hashemi, Milad and Duvenaud, David and Maddison, Chris},
  booktitle={International Conference on Machine Learning},
  pages={3831--3841},
  year={2021}
}

@inproceedings{hollmann2023tabpfn,
  title={Tab{PFN}: A transformer that solves small tabular classification problems in a second},
  author={Hollmann, Noah and M{\"u}ller, Samuel and Eggensperger, Katharina and Hutter, Frank},
  booktitle={International Conference on Learning Representations},
  year={2023}
}

@article{ma2024tabpfgen,
  title={Tab{PFG}en -- Tabular Data Generation with Tab{PFN}},
  author={Ma, Junwei and Dankar, Apoorv and Stein, George and Yu, Guangwei and Caterini, Anthony},
  journal={arXiv preprint arXiv:2406.05216},
  year={2024}
}

@inproceedings{shi2024tabdiff,
  title={Tab{D}iff: a Mixed-type Diffusion Model for Tabular Data Generation},
  author={Shi, Juntong and Xu, Minkai and Hua, Harper and Zhang, Hengrui and Ermon, Stefano and Leskovec, Jure},
  booktitle={International Conference on Learning Representations},
  year={2025}
}

@article{lecun2006tutorial,
  title={A tutorial on energy-based learning},
  author={LeCun, Yann and Chopra, Sumit and Hadsell, Raia and Ranzato, Marc'Aurelio and Huang, Fu Jie},
  journal={Predicting Structured Data},
  volume={1},
  year={2006}
}

@inproceedings{du2019implicit,
  title={Implicit generation and modeling with energy based models},
  author={Du, Yilun and Mordatch, Igor},
  booktitle={Advances in Neural Information Processing Systems},
  volume={32},
  year={2019}
}

@article{song2021train,
  title={How to train your energy-based models},
  author={Song, Yang and Kingma, Diederik P},
  journal={arXiv preprint arXiv:2101.03288},
  year={2021}
}

@inproceedings{grathwohl2020no,
  title={No {MCMC} for me: Amortized sampling for fast and stable training of energy-based models},
  author={Grathwohl, Will and Kelly, Jacob Jin and Hashemi, Milad and Norouzi, Mohammad and Swersky, Kevin and Duvenaud, David},
  booktitle={International Conference on Learning Representations},
  year={2021}
}

@inproceedings{zhang2024tabsyn,
  title={Mixed-Type Tabular Data Synthesis with Score-based Diffusion in Latent Space},
  author={Zhang, Hengrui and Zhang, Jiani and Srinivasan, Balasubramaniam and Shen, Zhengyuan and Qin, Xiao and Faloutsos, Christos and Rangwala, Huzefa and Karypis, George},
  booktitle={International Conference on Learning Representations},
  year={2024}
}

@inproceedings{li2023diffusion,
  title={Your Diffusion Model is Secretly a Zero-Shot Classifier},
  author={Li, Alexander C. and Prabhudesai, Mihir and Duggal, Shivam and Brown, Ellis and Pathak, Deepak},
  booktitle={Proceedings of the IEEE/CVF International Conference on Computer Vision},
  year={2023}
}

@inproceedings{schroder2024heat,
  title={Energy-Based Modelling for Discrete and Mixed Data via Heat Equations on Structured Spaces},
  author={Schr{\"o}der, Tobias and Ou, Zijing and Li, Yingzhen and Duncan, Andrew B.},
  booktitle={Advances in Neural Information Processing Systems},
  volume={37},
  year={2024},
  note={arXiv:2412.01019}
}

@inproceedings{chen2016xgboost,
  title={{XGBoost}: A Scalable Tree Boosting System},
  author={Chen, Tianqi and Guestrin, Carlos},
  booktitle={Proceedings of the 22nd ACM SIGKDD International Conference on Knowledge Discovery and Data Mining},
  year={2016}
}

@inproceedings{bischl2021openml,
  title={{OpenML} Benchmarking Suites},
  author={Bischl, Bernd and Casalicchio, Giuseppe and Feurer, Matthias and Gijsbers, Pieter and Hutter, Frank and Lang, Michel and Mantovani, Rafael G. and van Rijn, Jan N. and Vanschoren, Joaquin},
  booktitle={Proceedings of the Neural Information Processing Systems Track on Datasets and Benchmarks},
  year={2021}
}

@inproceedings{guo2017calibration,
  title={On Calibration of Modern Neural Networks},
  author={Guo, Chuan and Pleiss, Geoff and Sun, Yu and Weinberger, Kilian Q.},
  booktitle={International Conference on Machine Learning},
  year={2017}
}

@article{scassola2026sobering,
  title={A Sobering Look at Tabular Data Generation via Probabilistic Circuits},
  author={Scassola, Davide and Ponsford, Dylan and Javaloy, Adri{\'a}n and Saccani, Sebastiano and Bortolussi, Luca and Gouk, Henry and Vergari, Antonio},
  journal={arXiv preprint arXiv:2603.23016},
  year={2026}
}

@article{choi2020probabilistic,
  title={Probabilistic Circuits: A Unifying Framework for Tractable Probabilistic Models},
  author={Choi, YooJung and Vergari, Antonio and Van den Broeck, Guy},
  journal={Technical report, University of California, Los Angeles},
  year={2020}
}

@article{chow1968approximating,
  title={Approximating discrete probability distributions with dependence trees},
  author={Chow, C. K. and Liu, C. N.},
  journal={IEEE Transactions on Information Theory},
  volume={14},
  number={3},
  pages={462--467},
  year={1968}
}

@inproceedings{liu2021tractable,
  title={Tractable Regularization of Probabilistic Circuits},
  author={Liu, Anji and Van den Broeck, Guy},
  booktitle={Advances in Neural Information Processing Systems},
  year={2021}
}

@inproceedings{xu2019modeling,
  title={Modeling Tabular Data using Conditional {GAN}},
  author={Xu, Lei and Skoularidou, Maria and Cuesta-Infante, Alfredo and Veeramachaneni, Kalyan},
  booktitle={Advances in Neural Information Processing Systems},
  year={2019}
}

@inproceedings{borisov2023language,
  title={Language Models are Realistic Tabular Data Generators},
  author={Borisov, Vadim and Se{\ss}ler, Kathrin and Leemann, Tobias and Pawelczyk, Martin and Kasneci, Gjergji},
  booktitle={International Conference on Learning Representations},
  year={2023}
}

@article{hollmann2025accurate,
  title   = {Accurate predictions on small data with a tabular foundation model},
  author  = {Hollmann, Noah and M{\"u}ller, Samuel and Purucker, Lennart and Krishnakumar, Arjun and K{\"o}rfer, Max and Hoo, Shi Bin and Schirrmeister, Robin Tibor and Hutter, Frank},
  journal = {Nature},
  volume  = {637},
  number  = {8045},
  pages   = {319--326},
  year    = {2025},
  doi     = {10.1038/s41586-024-08328-6}
}

@inproceedings{muller2022transformers,
  title     = {Transformers Can Do {B}ayesian Inference},
  author    = {M{\"u}ller, Samuel and Hollmann, Noah and Pineda Arango, Sebastian and Grabocka, Josif and Hutter, Frank},
  booktitle = {International Conference on Learning Representations},
  year      = {2022}
}

@inproceedings{brown2020language,
  title     = {Language Models are Few-Shot Learners},
  author    = {Brown, Tom and Mann, Benjamin and Ryder, Nick and Subbiah, Melanie and Kaplan, Jared D. and Dhariwal, Prafulla and Neelakantan, Arvind and Shyam, Pranav and Sastry, Girish and Askell, Amanda and others},
  booktitle = {Advances in Neural Information Processing Systems},
  volume    = {33},
  pages     = {1877--1901},
  year      = {2020}
}

@inproceedings{sohl2015deep,
  title     = {Deep Unsupervised Learning using Nonequilibrium Thermodynamics},
  author    = {Sohl-Dickstein, Jascha and Weiss, Eric and Maheswaranathan, Niru and Ganguli, Surya},
  booktitle = {International Conference on Machine Learning},
  pages     = {2256--2265},
  year      = {2015}
}

@inproceedings{ho2020denoising,
  title     = {Denoising Diffusion Probabilistic Models},
  author    = {Ho, Jonathan and Jain, Ajay and Abbeel, Pieter},
  booktitle = {Advances in Neural Information Processing Systems},
  volume    = {33},
  pages     = {6840--6851},
  year      = {2020}
}

@inproceedings{song2021score,
  title     = {Score-Based Generative Modeling through Stochastic Differential Equations},
  author    = {Song, Yang and Sohl-Dickstein, Jascha and Kingma, Diederik P. and Kumar, Abhishek and Ermon, Stefano and Poole, Ben},
  booktitle = {International Conference on Learning Representations},
  year      = {2021}
}

@inproceedings{karras2022elucidating,
  title     = {Elucidating the Design Space of Diffusion-Based Generative Models},
  author    = {Karras, Tero and Aittala, Miika and Aila, Timo and Laine, Samuli},
  booktitle = {Advances in Neural Information Processing Systems},
  volume    = {35},
  year      = {2022}
}

@inproceedings{hoogeboom2021argmax,
  title     = {Argmax Flows and Multinomial Diffusion: Learning Categorical Distributions},
  author    = {Hoogeboom, Emiel and Nielsen, Didrik and Jaini, Priyank and Forr{\'e}, Patrick and Welling, Max},
  booktitle = {Advances in Neural Information Processing Systems},
  volume    = {34},
  year      = {2021}
}

@inproceedings{kotelnikov2023tabddpm,
  title     = {{TabDDPM}: Modelling Tabular Data with Diffusion Models},
  author    = {Kotelnikov, Akim and Baranchuk, Dmitry and Rubachev, Ivan and Babenko, Artem},
  booktitle = {International Conference on Machine Learning},
  pages     = {17564--17579},
  year      = {2023}
}

@inproceedings{poon2011sum,
  title     = {Sum-Product Networks: A New Deep Architecture},
  author    = {Poon, Hoifung and Domingos, Pedro},
  booktitle = {Conference on Uncertainty in Artificial Intelligence},
  pages     = {337--346},
  year      = {2011}
}

@inproceedings{peharz2020einsum,
  title     = {Einsum Networks: Fast and Scalable Learning of Tractable Probabilistic Circuits},
  author    = {Peharz, Robert and Lang, Steven and Vergari, Antonio and Stelzner, Karl and Molina, Alejandro and Trapp, Martin and Van den Broeck, Guy and Kersting, Kristian and Ghahramani, Zoubin},
  booktitle = {International Conference on Machine Learning},
  pages     = {7563--7574},
  year      = {2020}
}

@inproceedings{liu2023scaling,
  title     = {Scaling Up Probabilistic Circuits by Latent Variable Distillation},
  author    = {Liu, Anji and Zhang, Honghua and Van den Broeck, Guy},
  booktitle = {International Conference on Learning Representations},
  year      = {2023}
}

@inproceedings{vanbreugel2024position,
  title     = {Position: Why Tabular Foundation Models Should Be a Research Priority},
  author    = {van Breugel, Boris and van der Schaar, Mihaela},
  booktitle = {International Conference on Machine Learning},
  year      = {2024}
}

@article{aas2021explaining,
  title   = {Explaining individual predictions when features are dependent: More accurate approximations to {Shapley} values},
  author  = {Aas, Kjersti and Jullum, Martin and L{\o}land, Anders},
  journal = {Artificial Intelligence},
  volume  = {298},
  pages   = {103502},
  year    = {2021}
}

@article{redelmeier2024mcce,
  title   = {{MCCE}: {Monte Carlo} sampling of valid and realistic counterfactual explanations for tabular data},
  author  = {Redelmeier, Annabelle and Jullum, Martin and Aas, Kjersti and L{\o}land, Anders},
  journal = {Data Mining and Knowledge Discovery},
  year    = {2024}
}

@inproceedings{liu2020energy,
  title     = {Energy-based Out-of-distribution Detection},
  author    = {Liu, Weitang and Wang, Xiaoyun and Owens, John and Li, Yixuan},
  booktitle = {Advances in Neural Information Processing Systems},
  year      = {2020}
}

@article{cui2024tabular,
  title   = {Tabular Data Augmentation for Machine Learning: Progress and Prospects of Embracing Generative {AI}},
  author  = {Cui, Lingxi and Li, Huan and Chen, Ke and Shou, Lidan and Chen, Gang},
  journal = {arXiv preprint arXiv:2407.21523},
  year    = {2024}
}

@article{jager2021benchmark,
  title   = {A Benchmark for Data Imputation Methods},
  author  = {J{\"a}ger, Sebastian and Allhorn, Arndt and Bie{\ss}mann, Felix},
  journal = {Frontiers in Big Data},
  volume  = {4},
  pages   = {693674},
  year    = {2021},
  doi     = {10.3389/fdata.2021.693674}
}

@article{kingma2021variational,
  title={Variational diffusion models},
  author={Kingma, Diederik and Salimans, Tim and Poole, Ben and Ho, Jonathan},
  journal={Advances in neural information processing systems},
  volume={34},
  pages={21696--21707},
  year={2021}
}

@inproceedings{geifman2017selective,
  title     = {Selective Classification for Deep Neural Networks},
  author    = {Geifman, Yonatan and El-Yaniv, Ran},
  booktitle = {Advances in Neural Information Processing Systems},
  year      = {2017}
}

@article{brier1950verification,
  title={Verification of forecasts expressed in terms of probability},
  author={Brier, Glenn W.},
  journal={Monthly Weather Review},
  volume={78},
  number={1},
  pages={1--3},
  year={1950}
}

@article{kelly2021directly,
  title={Directly Training Joint Energy-Based Models for Conditional Synthesis and Calibrated Prediction of Multi-Attribute Data},
  author={Kelly, Jacob and Zemel, Richard and Grathwohl, Will},
  journal={arXiv preprint arXiv:2108.04227},
  year={2021}
}

@inproceedings{agarwal2021deep,
  title     = {Deep Reinforcement Learning at the Edge of the Statistical Precipice},
  author    = {Agarwal, Rishabh and Schwarzer, Max and Castro, Pablo Samuel and Courville, Aaron and Bellemare, Marc G.},
  booktitle = {Advances in Neural Information Processing Systems},
  volume    = {34},
  year      = {2021}
}

@inproceedings{sahoo2024simple,
  title={Simple and Effective Masked Diffusion Language Models},
  author={Sahoo, Subham Sekhar and Arriola, Marianne and Schiff, Yair and Gokaslan, Aaron and Marroquin, Edgar and Chiu, Justin T. and Rush, Alexander and Kuleshov, Volodymyr},
  booktitle={Advances in Neural Information Processing Systems},
  year={2024}
}

@inproceedings{shi2024simplified,
  title={Simplified and Generalized Masked Diffusion for Discrete Data},
  author={Shi, Jiaxin and Han, Kehang and Wang, Zhe and Doucet, Arnaud and Titsias, Michalis K.},
  booktitle={Advances in Neural Information Processing Systems},
  year={2024}
}

\appendix
\section{Cost accounting details}
\label{app:cost}

We report the cost of every method in FLOPs and in wall-clock time. 

We count the FLOPs of every run with our own counter which we will describe below.
% TODO(authors): true once the remaining counter runs (EBMs, MLP, PC, TabDiff, TabSyn, TabPFN at 2,048/4,096 rows) are done and the results are rebuilt from the counter only.

\subsection{Counting FLOPs}
\label{app:cost-measured}

We count all floating-point arithmetic that a run executes, in the forward and in the backward pass. Matrix products and attention count two FLOPs per multiply-accumulate. Every other floating-point operation counts one FLOP per output element: additions, multiplications, activations, exponentials and logarithms, normalisations and optimiser updates. Reductions, such as a sum over a dimension, count one FLOP per input element. We do not count data movement (copies, gathers, reshapes, concatenations and type casts), random-number generation, comparisons, integer arithmetic, or preprocessing outside the framework, such as the quantile transformation in scikit-learn. Every run has three phases, fitting, prediction and sampling, which we count separately. We note that elementwise operations make up less than $2\%$ of the FLOPs of every method, except the small PCs, whose sum layers are surrounded by exponentials and logarithms (about $9\%$ at 64 units).

\paragraph{JAX} The energy-based models, the MLP and the PC are written in JAX, and all their computation runs inside compiled (\texttt{jit}) functions. During a run, our counter wraps every compiled function. The first time a function is called with given input shapes, the counter traces it to JAX's intermediate representation (the jaxpr). The jaxpr contains the forward and the backward pass exactly as they are handed to the compiler, and the counter counts every operation in it at its full shape. It then adds this count each time the function is called. Loops are handled at two levels. A Python loop that calls a compiled function, such as the training loop over updates or the sampling loop over steps, is counted through the number of calls. A loop inside a compiled function (\texttt{scan}) is counted as its body times its number of iterations, and a conditional as its more expensive branch. Vectorised code, such as an operation mapped over a batch with \texttt{vmap}, appears in the jaxpr as one operation on the batched arrays, so it is counted at its full batched shape. Matrix products are the \texttt{dot\_general} and convolution operations of the jaxpr, at two FLOPs per multiply-accumulate. Elementwise operations are counted from the jaxpr's arithmetic primitives on floating-point arrays, at one FLOP per output element: addition, subtraction, multiplication, division, negation, absolute value, maximum, minimum, squares, reciprocals, powers, rounding, and the transcendental functions (exponential, logarithm, \texttt{log1p}, \texttt{tanh}, the logistic function, square roots, the error function and others). Reductions (sums, maxima, minima, products, cumulative sums and \texttt{argmax}) count one FLOP per input element, and scatter-adds one per added element. Comparisons, selections, sign functions, sorting and indexing are not counted.

\paragraph{PyTorch} The foundation models and the diffusion models run in PyTorch. Here the counter is a dispatch mode that sees every elementary (ATen) operation that actually executes, forward and backward, at its real shape. Because it records operations as they run, Python loops (over epochs, sampling steps or generated columns) are counted automatically, and a batched operation is counted with all of its elements: a batched matrix product counts the multiply-accumulates of every matrix in the batch. Fused kernels are split into their parts. Fused attention counts as its two matrix products, and a fused elementwise kernel, such as softmax or layer normalisation, counts as the elementary operations it consists of, so that both frameworks count the same arithmetic. Elementwise operations are recognised by PyTorch's own tags. Every operation tagged as pointwise (additions, multiplications, activations such as ReLU, GELU and SiLU, exponentials, logarithms, square roots and the like) counts one FLOP per floating-point output element. Every operation tagged as a reduction (sums, means, maxima and the like) counts one FLOP per input element. Fused kernels count a fixed number of FLOPs per input element, equal to the number of elementary operations they consist of: for example 5 for softmax, 4 for \texttt{logsumexp}, 8 for layer normalisation (14 in its backward pass) and 12 per parameter for the fused Adam update. Operations on integer tensors are not counted. Every operation of the reported runs falls into one of these groups. None of the models uses custom GPU kernels or \texttt{torch.compile}, which the counter would not see.

\paragraph{CPU} XGBoost runs as compiled C++ code on the CPU, outside both frameworks. We counted the floating-point instructions it executes with the opcode-mix tool of DynamoRIO, which counts every instruction as it is executed, so loops are included. Additions, subtractions, multiplications, divisions, minima, maxima and square roots count once per element. A vectorised (SIMD) instruction counts once per lane, and a fused multiply-add counts twice. Comparisons are not counted. On a CPU there is no separate class of elementwise operations: every arithmetic instruction counts, whether it is part of a gradient computation, a histogram sum or a split gain. Functions such as the exponential and the logarithm in XGBoost's loss are computed in software from these instructions, and are counted through them. We counted XGBoost~3.2.0 in this way on 37 synthetic tables ($500$--$8{,}000$ rows, $1$--$40$ features, $2$--$8$ classes, $0$--$50\%$ categorical features) with the benchmark's settings.
From these counts we fitted a cost model, which gives the cost of every XGBoost run. One fit costs
\[
  17.6\, S\, B_{\text{num}} + 69\, S\, B_{\text{cat}} + 1.0\, T\, n\, d \log_2(1 + S/T) + 22\, R\, o\, (n + n_{\text{val}})
\]
FLOPs, with $S$ split nodes and $T$ trees read off the fitted model, $R$ boosting rounds, $o$ outputs per round ($c$ classes, or $1$ for binary tasks), $B_{\text{num}}$ histogram bins summed over the numerical features and $B_{\text{cat}}$ levels summed over the categorical features. Prediction costs $n_{\text{test}}(1.6\,T + 4.2\,o)$. The model reproduces the counted FLOPs within $\pm13\%$ on the synthetic tables, and within $0.75$--$1.01\times$ on three held-out real tables.

% \subsection{Wall-clock time}
% Each run used one GH200 GPU and 8 CPU cores, except the column-by-column generators of TabPFN and TabICL and TabPFN's points at 2{,}048 and 4{,}096 context rows, which ran later on one A100. Warm-up is included: the first seed of each cell pays for JIT compilation (JAX) or for loading the checkpoint. Relative to the median of the other seeds, the first seed is $3.7\times$ slower for the MLP, $4.7\times$ for TabFM, $3.1\times$ for XGBoost, $2.2\times$ for JEM-ED, $2.0\times$ for JEM-SGLD and $1.1$--$1.2\times$ for the other methods. The median over five seeds absorbs most of this, but fixed overhead dominates for methods that take less than a second. The reported total is the sum of the per-phase medians.

\section{Reference rows}
\label{app:references}

\begin{table}[h]
  \centering
  \caption{Reference values that define $0$ and $1$ of the prediction and generation scores (interquartile mean over the suite's datasets of the per-dataset mean over 5 seeds). Class prior: predicts the training-split class frequencies $p(y)$ for every row (AUROC $=0.5$ by construction). Fully factorised: every column, the label included, sampled independently from its training-split marginal. Real data: a sample of training rows of the size of the test split.}
  \label{tab:references}
  \small
  \begin{tabular}{l|ccc|ccc|ccc}
\toprule
& \multicolumn{3}{c|}{Class prior} & \multicolumn{3}{c|}{Fully factorised} & \multicolumn{3}{c}{Real data} \\
Suite & Acc & Brier & NLL & wNMIS & C2ST-XGB & MLE & wNMIS & C2ST-XGB & MLE \\
\midrule
UCI-mixed & 0.729 & 0.388 & 0.576 & 0.804 & 0.019 & 0.505 & 0.994 & 0.996 & 0.865 \\
OpenML-CC18 & 0.572 & 0.522 & 0.828 & 0.891 & 0.226 & 0.520 & 0.976 & 0.971 & 0.865 \\
CC18 numerical & 0.549 & 0.534 & 0.891 & 0.844 & 0.111 & 0.523 & 0.974 & 0.973 & 0.891 \\
CC18 mixed & 0.619 & 0.495 & 0.791 & 0.941 & 0.604 & 0.511 & 0.980 & 0.965 & 0.741 \\
\bottomrule
\end{tabular}
\end{table}

Notably, the fully factorised sampler is not a weak floor on every metric: on the small, mostly categorical CC18-mixed tables it already reaches a C2ST-XGB of $0.60$ and a wNMIS of $0.94$, because these tables have few and weak dependencies, so the room between floor and ceiling is small there and the normalised scores of that subset are noisier. Its MLE is close to chance on every suite, as expected from labels that are independent of the features. The real-data MLE ($0.74$--$0.89$) is the AUROC that an XGBoost classifier reaches when trained on a real sample of the same size as the synthetic tables.

\begin{table}[h]
  \centering
  \caption{Headline scores with $95\%$ bootstrap intervals over datasets ($2{,}000$ resamples; interquartile mean of the per-dataset scores, at the best budgets of \Cref{tab:main}). With five datasets, the UCI-mixed intervals are wide.}
  \label{tab:scores-ci}
  \scriptsize
  \setlength{\tabcolsep}{3pt}
  \resizebox{\textwidth}{!}{\begin{tabular}{l|rccc|rccc}
\toprule
& \multicolumn{4}{c|}{UCI-mixed (5 datasets)} & \multicolumn{4}{c}{OpenML-CC18 (24 datasets)} \\
Method & Budget & Pred. & Gen. & Overall & Budget & Pred. & Gen. & Overall \\
\midrule
\multicolumn{9}{l}{\quad\textit{Energy-based models}} \\
JEM-ED & 2 & 0.476\,{\scriptsize[0.26, 0.67]} & 0.481\,{\scriptsize[0.33, 0.66]} & 0.478\,{\scriptsize[0.32, 0.64]} & 64 & 0.495\,{\scriptsize[0.25, 0.75]} & 0.610\,{\scriptsize[0.45, 0.74]} & 0.550\,{\scriptsize[0.39, 0.69]} \\
JEM-SGLD & 10 & 0.372\,{\scriptsize[0.19, 0.55]} & 0.434\,{\scriptsize[0.32, 0.57]} & 0.402\,{\scriptsize[0.25, 0.56]} & 10 & 0.544\,{\scriptsize[0.34, 0.74]} & 0.603\,{\scriptsize[0.49, 0.71]} & 0.573\,{\scriptsize[0.44, 0.69]} \\
TabED-joint & 2 & 0.419\,{\scriptsize[0.17, 0.63]} & 0.459\,{\scriptsize[0.26, 0.64]} & 0.438\,{\scriptsize[0.22, 0.61]} & 128 & 0.522\,{\scriptsize[0.32, 0.71]} & 0.522\,{\scriptsize[0.33, 0.73]} & 0.522\,{\scriptsize[0.38, 0.67]} \\
\multicolumn{9}{l}{\quad\textit{Tabular diffusion models}} \\
TabDiff & 2000 & 0.277\,{\scriptsize[0.11, 0.58]} & 0.858\,{\scriptsize[0.83, 0.88]} & 0.488\,{\scriptsize[0.30, 0.71]} & 8000 & 0.447\,{\scriptsize[0.28, 0.64]} & 0.690\,{\scriptsize[0.53, 0.85]} & 0.555\,{\scriptsize[0.44, 0.69]} \\
TabSyn & 1250 & 0.415\,{\scriptsize[0.18, 0.65]} & 0.861\,{\scriptsize[0.76, 0.92]} & 0.598\,{\scriptsize[0.37, 0.77]} & 1250 & 0.533\,{\scriptsize[0.31, 0.76]} & 0.843\,{\scriptsize[0.78, 0.90]} & 0.670\,{\scriptsize[0.51, 0.79]} \\
\multicolumn{9}{l}{\quad\textit{Tabular foundation models}} \\
TabPFN & 4096 & 0.476\,{\scriptsize[0.25, 0.67]} & 0.813\,{\scriptsize[0.75, 0.90]} & 0.622\,{\scriptsize[0.43, 0.77]} & 4096 & 0.636\,{\scriptsize[0.46, 0.79]} & 0.920\,{\scriptsize[0.88, 0.95]} & 0.765\,{\scriptsize[0.65, 0.86]} \\
TabICL & 1 & 0.497\,{\scriptsize[0.29, 0.70]} & 0.836\,{\scriptsize[0.77, 0.89]} & 0.644\,{\scriptsize[0.47, 0.79]} & 16 & 0.668\,{\scriptsize[0.48, 0.83]} & 0.934\,{\scriptsize[0.88, 0.96]} & 0.790\,{\scriptsize[0.66, 0.89]} \\
\multicolumn{9}{l}{\quad\textit{Probabilistic circuits}} \\
PC (HCLT) & 4096 & 0.458\,{\scriptsize[0.25, 0.62]} & 0.870\,{\scriptsize[0.79, 0.90]} & 0.631\,{\scriptsize[0.46, 0.74]} & 64 & 0.416\,{\scriptsize[0.26, 0.62]} & 0.676\,{\scriptsize[0.61, 0.76]} & 0.531\,{\scriptsize[0.41, 0.67]} \\
\multicolumn{9}{l}{\quad\textit{Single-task references}} \\
MLP (CE only) & -- & 0.461\,{\scriptsize[0.27, 0.64]} & -- & -- & -- & 0.483\,{\scriptsize[0.22, 0.74]} & -- & -- \\
XGBoost & 25 & 0.487\,{\scriptsize[0.27, 0.65]} & -- & -- & 500 & 0.597\,{\scriptsize[0.41, 0.77]} & -- & -- \\
TabFM & 32 & 0.511\,{\scriptsize[0.30, 0.72]} & -- & -- & 4 & 0.684\,{\scriptsize[0.49, 0.85]} & -- & -- \\
\bottomrule
\end{tabular}}
\end{table}

\section{Energy-based samplers for the foundation models}
\label{app:sampler-ablation}
This section shows an ablation study where we compare the chain rule based sampler for foundation models with energy based sampling as proposed in \citep{ma2024tabpfgen}. Table \Cref{tab:sampler-ablation} compares generative performance of these 2 sampling strategies for TabPFN and TabICL.
\begin{table}[h]
  \centering
  \caption{Generation from TabPFN and TabICL with their own column-by-column generators (used in the main text) and with energy-based samplers that read the classifier's logits as a class-conditional energy, at the best budget.}
  \label{tab:sampler-ablation}
  \scriptsize
  \begin{tabular}{lll|cccc|rr}
\toprule
Suite & Model & Sampler & wNMIS & C2ST-XGB & MLE & Gen. & sample TFLOP & sample s \\
\midrule
UCI-mixed & TabPFN@1024 & column-by-column & 0.979 & 0.547 & 0.837 & 0.783 & 297 & 34.9 \\
 & TabPFN@1024 & TabPFGen (energy, prior start) & 0.814 & 0.001 & 0.753 & 0.241 & 4{,}080 & 268 \\
 & TabICL@16 & column-by-column & 0.981 & 0.645 & 0.840 & 0.825 & 305 & 17.8 \\
 & TabICL@16 & TabICL-Gen (energy, prior start) & 0.811 & 0.001 & 0.707 & 0.230 & $3.3{\times}10^{4}$ & 4{,}293 \\
\midrule
OpenML-CC18 & TabPFN@1024 & column-by-column & 0.970 & 0.881 & 0.846 & 0.911 & 17.1 & 12.5 \\
 & TabPFN@1024 & TabPFGen (energy, prior start) & 0.904 & 0.104 & 0.799 & 0.279 & 58.3 & 28.3 \\
 & TabICL@16 & column-by-column & 0.968 & 0.883 & 0.833 & 0.934 & 15 & 4.15 \\
 & TabICL@16 & TabICL-Gen (energy, prior start) & 0.906 & 0.181 & 0.815 & 0.352 & 895 & 75.2 \\
\bottomrule
\end{tabular}
\end{table}

TabPFGen \citep{ma2024tabpfgen} reads TabPFN's logits as a class-conditional energy $E(\vx \mid y) = -f(\vx)[y]$ and samples it with SGLD (step size $0.01$, noise $0.1$, with its best-batch selection), using a prompt of 500 training rows. We start its chains from the same prior as the JEM samplers (standard normal numerical features, uniformly drawn category levels) instead of from noised training rows, so that no sample is seeded with real records, and resample every categorical feature exactly from its enumerated conditional every 10 steps (Metropolis-within-Gibbs for features with at least 50 levels). TabICL-Gen applies the same sampler and settings to a differentiable re-implementation of TabICL's forward pass (verified against the package's predictions to $2\cdot10^{-6}$), with 2 ensemble members, a prompt of at most 2{,}048 rows and 200 steps.

We find that neither of the two energy-based samplers seems to pull a chain from the prior towards the data manifold within the sampling budget: C2ST-XGB is $0.001$ on UCI-mixed and $0.10$--$0.18$ on CC18, and the generation score is $0.23$--$0.35$, against $0.78$--$0.93$ for the column-by-column generators of the same models, which also cost $3$--$108\times$ fewer sampling FLOPs. We note that when the chains are started from noised training rows, as published in \citep{ma2024tabpfgen}, TabPFGen reached a C2ST-XGB of $0.63$ on CC18 in an earlier run of our benchmark, but its samples are then very close to real records from the training data, which is not what we desire from a generator and would be an unfair comparison to other methods in this study.

\section{Datasets}
\label{app:datasets}

\paragraph{UCI-mixed} \Cref{tab:uci-datasets} lists the five UCI classification datasets with their feature types, as used by our loader. Four of them contain numerical and categorical features. MAGIC is purely numerical. For Diabetes, the target is binarised (readmitted or not) and the 26 categorical features with the most levels are kept. The training splits (64\% of the rows) range from 7{,}891 rows (Online Shoppers) to 65{,}130 rows (Diabetes).

\begin{table}[h]
  \centering
  \caption{UCI datasets used in this work. \#num / \#cat: numerical / categorical input features (target excluded). Type: which feature types the dataset contains.}
  \label{tab:uci-datasets}
  \small
  \begin{tabular}{lrrrrrl}
    \toprule
    Dataset & UCI ID & Rows & \#num & \#cat & Classes & Type \\
    \midrule
    Adult & 2 & 48{,}842 & 6 & 8 & 2 & both \\
    Default of credit card clients & 350 & 30{,}000 & 14 & 9 & 2 & both \\
    Diabetes 130-US hospitals & 296 & 101{,}766 & 9 & 26 & 2 & both \\
    MAGIC gamma telescope & 159 & 19{,}020 & 10 & 0 & 2 & numerical \\
    Online Shoppers purchasing intention & 468 & 12{,}330 & 10 & 7 & 2 & both \\
    \bottomrule
  \end{tabular}
\end{table}

\paragraph{OpenML-CC18} \Cref{tab:cc18-datasets} lists the 24 CC18 datasets with their feature types, as recorded in OpenML's feature metadata and used by our loader. The 18 numerical datasets are the purely numerical selection of \citet{ma2024tabpfgen}. The remaining 6 datasets are all CC18 datasets with categorical features that satisfy the same constraints ($\le 2{,}000$ rows, $\le 100$ features, $\le 10$ classes, no missing values). Three of them (cmc, credit-g, ilpd) contain both numerical and categorical features, and three (analcatdata\_dmft, car, tic-tac-toe) are purely categorical. Following \Cref{sec:background}, we call this subset \emph{mixed-type}. Integer-valued columns that OpenML types as numeric (e.g.\ the 1--5 attributes of balance-scale and the word counts of analcatdata\_authorship) are treated as numerical.

\begin{table}[h]
  \centering
  \caption{OpenML-CC18 datasets used in this work. \#num / \#cat: numerical / categorical input features (target excluded). Type: which feature types the dataset contains.}
  \label{tab:cc18-datasets}
  \small
  \begin{tabular}{lrrrrrl}
    \toprule
    Dataset & OpenML ID & Rows & \#num & \#cat & Classes & Type \\
\midrule
\multicolumn{7}{l}{\textit{Numerical subset (18)}} \\
analcatdata\_authorship & 458 & 841 & 70 & 0 & 4 & numerical \\
balance-scale & 11 & 625 & 4 & 0 & 3 & numerical \\
banknote-authentication & 1462 & 1372 & 4 & 0 & 2 & numerical \\
blood-transfusion-service-center & 1464 & 748 & 4 & 0 & 2 & numerical \\
climate-model-simulation-crashes & 40994 & 540 & 20 & 0 & 2 & numerical \\
diabetes (Pima) & 37 & 768 & 8 & 0 & 2 & numerical \\
kc2 & 1063 & 522 & 21 & 0 & 2 & numerical \\
mfeat-fourier & 14 & 2000 & 76 & 0 & 10 & numerical \\
mfeat-karhunen & 16 & 2000 & 64 & 0 & 10 & numerical \\
mfeat-morphological & 18 & 2000 & 6 & 0 & 10 & numerical \\
mfeat-zernike & 22 & 2000 & 47 & 0 & 10 & numerical \\
pc1 & 1068 & 1109 & 21 & 0 & 2 & numerical \\
pc3 & 1050 & 1563 & 37 & 0 & 2 & numerical \\
pc4 & 1049 & 1458 & 37 & 0 & 2 & numerical \\
qsar-biodeg & 1494 & 1055 & 41 & 0 & 2 & numerical \\
steel-plates-fault & 40982 & 1941 & 27 & 0 & 7 & numerical \\
vehicle & 54 & 846 & 18 & 0 & 4 & numerical \\
wdbc & 1510 & 569 & 30 & 0 & 2 & numerical \\
\midrule
\multicolumn{7}{l}{\textit{Mixed-type subset (6): numerical and categorical features}} \\
cmc & 23 & 1473 & 2 & 7 & 3 & both \\
credit-g & 31 & 1000 & 7 & 13 & 2 & both \\
ilpd & 1480 & 583 & 9 & 1 & 2 & both \\
\multicolumn{7}{l}{\textit{\quad purely categorical features}} \\
analcatdata\_dmft & 469 & 797 & 0 & 4 & 6 & categorical \\
car & 40975 & 1728 & 0 & 6 & 4 & categorical \\
tic-tac-toe & 50 & 958 & 0 & 9 & 2 & categorical \\
    \bottomrule
  \end{tabular}
\end{table}

\section{Hyperparameters and implementation details}
\label{app:hyperparameters}
\label{app:budgets}

No method is tuned per dataset or per suite. Every method is swept along one hyperparameter that controls its computational budget (its \emph{budget hyperparameter}, \Cref{tab:budgets}). All of its other hyperparameters are held fixed across suites, datasets, seeds and budgets, at the values listed below. For published methods these are the authors' published settings or the package defaults, unless stated otherwise. The hyperparameters of the energy-based models were set once and are shared by all suites.
% NOTE(authors): state on which data/splits the JEM settings (w, alpha, t_num, lr, width) were chosen in the 2026-09-07 retune.

\subsection{Budget grids}

\begin{table}[h]
  \centering
  \caption{Budget hyperparameter and grid per method; the grids are the same on every suite. The headline (default) budget is underlined; \Cref{tab:main} instead reports each method at its best budget.}
  \label{tab:budgets}
  \small
  \begin{tabular}{lll}
    \toprule
    Method & Budget hyperparameter & Grid \\
    \midrule
    JEM-ED & ED particles $M$ & 1, 2, 4, 8, 16, \underline{32}, 64, 128 \\
    JEM-SGLD & SGLD steps per update $K$ & 2, 5, 10, \underline{20}, 40 \\
    TabED-joint & ED particles $M$ & 1, 2, 4, 8, 16, \underline{32}, 64, 128 \\
    TabPFN & context rows (classifier and generator) & 64, 128, 256, 512, \underline{1024}, 2048, 4096 \\
    TabICL & ensemble members (classifier and generator) & 1, 2, 4, 8, \underline{16} \\
    TabDiff & training epochs & 500, 1000, 2000, 4000, \underline{8000} \\
    TabSyn & latent-diffusion epochs (VAE fixed at 4000) & 625, 1250, 2500, 5000, \underline{10000} \\
    PC (HCLT) & units per variable $K_{\text{PC}}$ & 64, 256, \underline{1024}, 2048, 4096 \\
    \midrule
    MLP & -- & -- \\
    XGBoost & maximum boosting rounds & 5, 10, 25, 100, \underline{500} \\
    TabFM & ensemble members & 1, 2, 4, 8, \underline{32} \\
    \bottomrule
  \end{tabular}
\end{table}

The \emph{headline budget} is the published or default value where it lies on the grid and is underlined in \cref{tab:budgets}; it is used in the sampler ablation (\Cref{app:sampler-ablation}), whereas \cref{tab:main} reports each method at its best budget. 
%$M=32$ and $K=20$ (our defaults for the energy-based models), 500 boosting rounds, 32 TabFM members, 8{,}000 TabDiff epochs and 10{,}000 TabSyn epochs. TabPFN uses its largest swept context (1{,}024 rows) and TabICL its largest ensemble (16 members, twice its package default of 8). For the PC we use $K_{\text{PC}} = 1{,}024$, the middle of the values tuned by \citet{scassola2026sobering}. Three budget hyperparameters stop changing the computation at their upper end. XGBoost's early stopping usually ends training below 25 rounds. TabSyn's early stopping ends latent-diffusion training before about 2{,}500 epochs (\Cref{app:cost}). TabICL's ensemble size changes its generator's cost but not its fidelity.

\subsection{Software}

The energy-based models, the MLP and the PC are our own implementations in JAX~0.6.2 with Equinox~0.13.6 and Optax~0.2.8. The foundation models use their official packages: \texttt{tabpfn}~2.0.3 with the TabPFN v2 classifier and regressor weights, \texttt{tabicl}~2.2.0 with its default checkpoint, and \texttt{tabfm}~1.0.1 with its PyTorch backend and v1.0.0 weights. XGBoost is \texttt{xgboost}~3.2.0, and preprocessing uses scikit-learn~1.6.1. For TabDiff and TabSyn we vendor the model code from the authors' repositories (TabDiff: commit \texttt{5ecdb33}, MIT licence; TabSyn: commit \texttt{cb5ac0f}, Apache-2.0) and drive it with our own training loop, which reproduces the upstream training scripts (optimiser, schedules, EMA and checkpoint selection) on our preprocessed data. All runs use Python~3.11.
% NOTE(authors): confirm the JAX/PyTorch versions against the Helios environment (versions above from the project lock for Linux/Python 3.11).

\subsection{Energy-based models and MLP}

\paragraph{Network architecture and training} JEM-ED, JEM-SGLD, TabED-joint and the MLP share one network, which follows the tabular EBM of \citet{schroder2023training}. Each categorical column is one-hot encoded and mapped by its own learned linear embedding to 4 dimensions. The embeddings are concatenated with the quantile-normalised numerical features and fed to a fully connected network with two hidden layers of 512 units, Swish activations and no normalisation layers. For JEM-ED, JEM-SGLD and the MLP, the output layer gives one logit per class, $f_\theta(\vx)$, with energy $E_\theta(\vx) = -\log\sum_y \exp f_\theta(\vx)[y]$. TabED-joint receives the label as an additional categorical input column, with its own embedding, and has a single output, the joint energy $E_\theta(\vx, y)$. All four methods train for 2{,}000 updates on minibatches of 256 rows, drawn uniformly with replacement from the training split. None of them uses early stopping or the validation split.

\paragraph{JEM-ED} We optimise with Adam with learning rate $5\cdot10^{-4}$. The objective is \eqref{eq:jem-ed-objective} with $\alpha = 0.6$ on the cross-entropy term and weight 1 on the ED term. The perturbation kernel \eqref{eq:mixed-kernel} uses $t_{\text{num}} = 0.2$, the standard deviation of the Gaussian noise on numerical features, and $t_{\text{cat}} = 0.1$, which keeps a categorical value with probability $e^{-0.1} \approx 0.90$. The offset is $w = 0.5$. It is added to the \emph{mean} over the $M$ particles (\eqref{eq:ed-tab}), so it is the same stabilisation at every $M$; in the convention of \citet{schroder2024heat}, who add it to the sum, it corresponds to $wM$. The budget hyperparameter is the number of particles $M$.

\paragraph{JEM-SGLD} Training follows \citet{grathwohl2019your}, using Adam and their learning rate of $10^{-4}$. It uses persistent contrastive divergence with a replay buffer of 1{,}000 particles. Every update draws one chain per example of the minibatch from the buffer and restarts 5\% of the drawn chains from the common prior (\Cref{sec:cand-ebm}). Each chain then runs $K$ steps of the Langevin-within-Gibbs sampler and is written back to the buffer. 
Each step is an SGLD step on the numerical features, $\vx \leftarrow \vx - \tfrac{\epsilon}{2}\nabla_{\vx}E_\theta + \sqrt{\epsilon}\,\bm{\xi}$ with $\epsilon = 0.01$ (noise standard deviation $0.1$), followed by an exact Gibbs sweep over the categorical columns. We depart from \citet{grathwohl2019your} in three ways: a buffer of 1{,}000 instead of 10{,}000 particles; a Langevin step whose noise matches its step size instead of their step size 1 with noise standard deviation $0.01$ (following \citep{nijkamp2020anatomy}); and the Gibbs sweep. The budget hyperparameter is the number of steps per update $K$.

\paragraph{TabED-joint} TabED-joint adopts the hyperparameters from JEM-ED: Adam with learning rate $5\cdot10^{-4}$, $t_{\text{num}} = 0.2$, $t_{\text{cat}} = 0.1$, $w = 0.5$ and $M$ particles as budget hyperparameter. The objective is ED on the joint energy $E_\theta(\vx, y)$ alone, without a cross-entropy term. The label column is perturbed by the same uniform kernel as the other categorical columns. The classifier is $p_\theta(y \mid \vx) \propto \exp(-E_\theta(\vx, y))$.

\paragraph{MLP} The MLP is trained with the cross-entropy loss only, using Adam with learning rate $10^{-3}$. It has no budget hyperparameter.

\paragraph{Generation} All three energy-based models generate with the same sampler (\Cref{sec:cand-ebm}): 200 Langevin-within-Gibbs steps with SGLD step size $0.01$ and an exact Gibbs sweep over all levels of every categorical column per step, from the common prior. Each row's label is then drawn from the model's classifier $p_\theta(y \mid \vx)$. For TabED-joint, the sampler runs on the marginal energy $-\log \sum_{c} \exp(-E_\theta(\vx, c))$, so every energy evaluation costs one forward pass per class.

\subsection{Tabular foundation models}

All foundation models are used with their publicly available pretrained weights.

\paragraph{TabPFN} We use the official implementation of TabPFN, the \texttt{tabpfn} package (version 2.0.3)\footnote{\url{https://github.com/PriorLabs/TabPFN}}, with its pretrained TabPFN v2 weights. The package provides a classifier (\texttt{TabPFNClassifier}) and a regressor (\texttt{TabPFNRegressor}). These are two separate models, each pretrained for its own task with its own weights. The classifier half of the joint model uses only the classifier; the generator uses both.

Both halves are conditioned on a context of training rows, drawn uniformly at random without replacement from the training split, separately for each seed. The context size is the budget hyperparameter (64 to 4{,}096 rows, \Cref{tab:budgets}). If the training split is smaller, the whole split is used, which is the case for all 24 CC18 datasets at 2{,}048 and 4{,}096 rows, for 18 of them at 1{,}024 rows and for 8 at 512 rows; the UCI training splits are larger than every budget. The points at 2{,}048 and 4{,}096 rows were run later, on one A100 (\Cref{app:cost}). The classifier and the generator use the same context size, but draw their rows independently. The generator's context contains the label as an additional column.

The \emph{classifier} is a single forward pass of the pretrained classifier with one ensemble member, on inputs standardised with the mean and standard deviation of the context. It bypasses the package's own preprocessing and ensembling, and its class probabilities are the softmax of the logits. This differentiable forward pass is the one on which TabPFGen builds \citep{ma2024tabpfgen}, and its logits also define the energy in the sampler ablation (\Cref{app:sampler-ablation}). It is cheaper than the package's default prediction, which averages several ensemble members over its own preprocessing, and may be less accurate.

The \emph{generator} is TabPFN's chain-rule construction (\eqref{eq:tfm-chain}; \citealp{hollmann2025accurate}), in our implementation (\Cref{sec:cand-tfm}). Unlike the classifier, it calls \texttt{TabPFNClassifier} and \texttt{TabPFNRegressor} through the package's standard interface, with their own preprocessing and 8 ensemble members each. It generates as many rows as the test split has, one column at a time, in schema order with the label last:
\begin{itemize}
  \item The first column is drawn from its empirical distribution in the context. A column that is constant in the context is copied.
  \item For every later column $j$, a model is fitted on the context, with columns $1, \dots, j-1$ as inputs and column $j$ as target: the regressor for numerical columns and the classifier for categorical columns, including the label. It then predicts column $j$ for the rows generated so far, and one value per row is drawn from its predictive distribution at temperature 1. For the regressor, this is TabPFN's discretised distribution over target values.
  \item TabPFN v2 accepts at most 10 classes. Categorical columns with more levels are sampled hierarchically: first one of at most 10 balanced groups of levels, then a level within that group.
\end{itemize}

\paragraph{TabICL} The \emph{classifier} is \texttt{TabICLClassifier} with the whole training split as its context and the number of ensemble members given by the budget hyperparameter. Categorical columns are passed as categorical columns of a data frame, so that TabICL applies its own categorical preprocessing. The \emph{generator} is the package's \texttt{TabICLUnsupervised} with the same number of ensemble members, fitted on a random subsample of at most 2{,}000 rows of the joint table (features and label). It generates the columns in schema order with the label last, at temperature 1.

\paragraph{TabFM} TabFM is a single-task classifier in this study \citep{kong2026tabfm}. It runs with the PyTorch backend in single precision (float32 instead of the package's bfloat16 default), with the training split as its context and the number of ensemble members given by the budget hyperparameter. Its default of 32 members is the headline budget.

\subsection{Tabular diffusion models}

\paragraph{TabDiff} TabDiff runs with the upstream architecture and training configuration \citep{shi2024tabdiff}: AdamW with learning rate $10^{-3}$ and no weight decay, batch size 4{,}096, a reduce-on-plateau schedule (patience 50 epochs, factor $0.9$) and an EMA of the weights with rate $0.997$, applied once per epoch. The weight of the categorical loss is annealed as upstream, and sampling uses 50 diffusion steps. Upstream keeps the best EMA checkpoint, but only considers checkpoints after epoch 4{,}000. We keep this rule, so the budgets of 500 to 2{,}000 epochs use the final EMA weights and the 4{,}000- and 8{,}000-epoch budgets the best EMA checkpoint.

\paragraph{TabSyn} TabSyn also runs with the upstream configuration \citep{zhang2024tabsyn}. The VAE has 2 transformer layers, token dimension 4, one attention head and factor 32, and is trained for 4{,}000 epochs with the $\beta$ schedule from $10^{-2}$ to $10^{-5}$ (factor $0.7$); its final-epoch weights are used. The latent-diffusion MLP has hidden width 1{,}024 and stops early after 500 epochs without improvement of its training loss. Both stages use Adam with learning rate $10^{-3}$, no weight decay and batch size 4{,}096. Sampling uses the 50-step EDM sampler.

\paragraph{Diffusion classifiers} Both models classify a row by Bayes' rule over their ELBOs (\Cref{sec:cand-diffusion}), estimated from 250 diffusion-time draws per row. The draws lie on a shared grid with a half-step offset and are the same for every class. For TabDiff, which keeps the label column clean and therefore scores $\log p(\vx \mid y)$, the training class prior is added. For TabSyn the label is inside the VAE latent, so the bound is joint and no prior is added.

\subsection{Probabilistic circuit}

The PC follows TabPC's HCLT recipe \citep{scassola2026sobering}. Its Chow--Liu tree is learned on the training split from pairwise mutual information, with numerical columns discretised into 16 quantile bins for this estimate only, and is rooted at its centre. Numerical columns have Gaussian leaves, with the standard deviation floored at $0.05$, and categorical columns, label included, have categorical leaves. Training uses RAdam with learning rate $0.1$ and batch size 512 for at most 200 epochs, with early stopping on the validation log-likelihood (patience 10) and the learning rate multiplied by $0.85$ when the validation log-likelihood plateaus. The classifier is the exact conditional $p(y \mid \vx) = p(\vx, y)/p(\vx)$, and samples are drawn by ancestral sampling.

\subsection{XGBoost}

XGBoost uses its native categorical splits and the package defaults otherwise (histogram trees, maximum depth 6, learning rate $0.3$). The budget hyperparameter is the maximum number of boosting rounds, and training stops early when the validation log-loss has not improved for 20 rounds.

\subsection{Reference rows}

FF, the real-data row and the class prior (\Cref{app:references}) have no hyperparameters.

% Full per-budget sweep tables (formerly app:sweeps), currently not included:
% The tables below list every method and budget for each suite. TFLOP: total FLOPs (fit $+$ predict $+$ sample, $10^{12}$); s: total wall-clock seconds. Pred.\ and Gen.: prediction and generation scores (\Cref{sec:experimental-setup}, reference values in \Cref{tab:references}); Overall: their geometric mean.
%\input{tables/sweep_uci7.tex}
%\input{tables/sweep_cc18.tex}
%\input{tables/sweep_cc18_numerical.tex}
%\input{tables/sweep_cc18_mixed.tex}

\section{Training energy-based joint models: energy discrepancy versus SGLD}
\label{sec:jem-cost}
\label{app:jem-training}
This appendix compares the two ways of training the energy-based joint models of \Cref{sec:cand-ebm} within one family: JEM-ED and JEM-SGLD share the backbone and the number of parameter updates, and TabED-joint applies ED to the full joint instead of the marginal.

\begin{table}[t]
  \centering
  \caption{JEM-ED and JEM-SGLD along their budget hyperparameters (number of ED particles $M$, number of SGLD steps per update $K$). Both share the backbone and the number of parameter updates. TFLOP: \emph{training} FLOPs ($10^{12}$); s: measured training wall-clock in seconds (median, GH200). Accuracy and C2ST-XGB are interquartile means over datasets of the per-dataset mean over 5 seeds.}
  \label{tab:jem-cost}
  \scriptsize
  \setlength{\tabcolsep}{3pt}
  \resizebox{\textwidth}{!}{\begin{tabular}{ll|ccrr|ccrr|ccrr}
\toprule
& & \multicolumn{4}{c|}{UCI-mixed} & \multicolumn{4}{c|}{CC18 numerical} & \multicolumn{4}{c}{CC18 mixed} \\
Method & Budget & Acc & C2ST-XGB & TFLOP & s & Acc & C2ST-XGB & TFLOP & s & Acc & C2ST-XGB & TFLOP & s \\
\midrule
JEM-ED & $M=1$ & 0.854 & 0.132 & 1.7 & 19 & 0.844 & 0.367 & 1.7 & 7 & 0.723 & 0.736 & 1.7 & 17 \\
 & $M=2$ & 0.854 & 0.160 & 2.6 & 22 & 0.848 & 0.394 & 2.6 & 7 & 0.726 & 0.829 & 2.6 & 17 \\
 & $M=4$ & 0.854 & 0.139 & 4.4 & 25 & 0.848 & 0.406 & 4.3 & 6 & 0.726 & 0.848 & 4.3 & 20 \\
 & $M=8$ & 0.854 & 0.118 & 7.8 & 36 & 0.847 & 0.412 & 7.7 & 7 & 0.729 & 0.843 & 7.7 & 29 \\
 & $M=16$ & 0.854 & 0.098 & 14.8 & 70 & 0.849 & 0.424 & 14.5 & 7 & 0.728 & 0.835 & 14.6 & 54 \\
 & $M=32$ & 0.854 & 0.088 & 28.7 & 170 & 0.848 & 0.412 & 28.0 & 7 & 0.724 & 0.847 & 28.3 & 131 \\
 & $M=64$ & 0.854 & 0.083 & 56.5 & 550 & 0.847 & 0.424 & 55.2 & 7 & 0.723 & 0.854 & 55.8 & 424 \\
 & $M=128$ & 0.854 & 0.078 & 112.1 & 1928 & 0.847 & 0.434 & 109.6 & 8 & 0.721 & 0.851 & 110.8 & 1516 \\
\midrule
JEM-SGLD & $K=2$ & 0.826 & 0.037 & 48.6 & 32 & 0.834 & 0.080 & 2.8 & 13 & 0.732 & 0.729 & 15.2 & 26 \\
 & $K=5$ & 0.828 & 0.100 & 118.7 & 35 & 0.841 & 0.236 & 4.6 & 13 & 0.719 & 0.804 & 35.3 & 27 \\
 & $K=10$ & 0.827 & 0.104 & 235.6 & 39 & 0.846 & 0.326 & 7.4 & 13 & 0.724 & 0.820 & 69.0 & 29 \\
 & $K=20$ & 0.829 & 0.106 & 469.9 & 47 & 0.847 & 0.430 & 13.1 & 13 & 0.720 & 0.879 & 136.2 & 33 \\
 & $K=40$ & 0.829 & 0.107 & 938.0 & 62 & 0.846 & 0.451 & 24.5 & 15 & 0.724 & 0.860 & 270.7 & 40 \\
\bottomrule
\end{tabular}}
\end{table}

\begin{figure}[t]
  \centering
  \includegraphics[width=\textwidth]{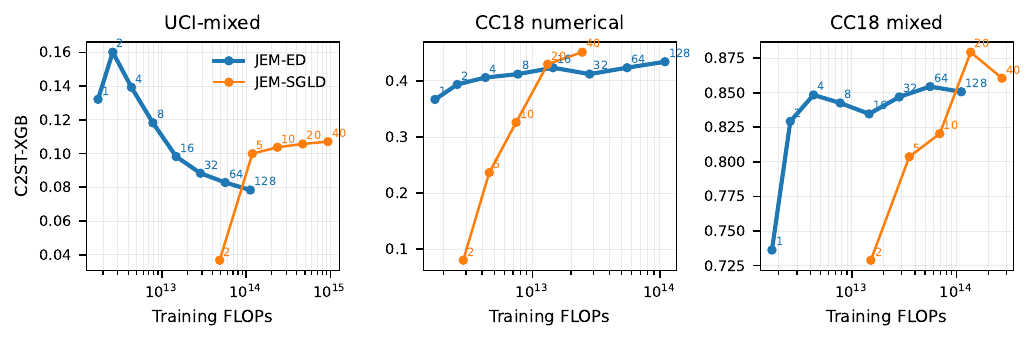}
  \caption{Joint fidelity (C2ST-XGB) against training FLOPs for JEM-ED (varying $M$) and JEM-SGLD (varying $K$) on the same backbone. Point labels give $M$ or $K$.}
  \label{fig:jem-cost}
\end{figure}

\paragraph{Scaling with the budget hyperparameter}
\Cref{tab:jem-cost} and \Cref{fig:jem-cost} trace both JEMs along their budget hyperparameters. Training FLOPs of JEM-ED grow linearly with $M$ and are nearly identical across UCI-mixed and CC18 ($1.7$ TFLOP at $M=1$ to $110$--$112$ TFLOP at $M=128$), because each update evaluates $M{+}1$ copies of the batch. Training FLOPs of JEM-SGLD grow linearly with $K$ but depend strongly on the data: on CC18-numerical they range from $2.8$ to $24.5$ TFLOP, whereas on the high-cardinality categorical features of UCI-mixed every Gibbs sweep costs one forward pass per category level, and the same range of $K$ spans $49$ to $938$ TFLOP.

\paragraph{The quality of JEM-ED is insensitive to $M$}
The most consequential observation in \Cref{tab:jem-cost} is that JEM-ED's accuracy is almost flat in $M$ and its fidelity saturates early: C2ST-XGB on CC18-numerical rises from $0.37$ at $M=1$ to $0.41$ at $M=8$ and stays between $0.41$ and $0.43$, and on CC18-mixed it stays between $0.84$ and $0.85$ from $M=4$. On UCI-mixed, fidelity is even highest at $M=2$ ($0.160$) and \emph{decreases} for $M \ge 4$; we do not yet have an explanation for this and treat it as an open issue. JEM-SGLD, in contrast, needs $K \ge 10$--$20$ steps before its fidelity is competitive: at $K=2$ its C2ST-XGB is $0.04$ on UCI-mixed and $0.08$ on CC18-numerical. Consequently, at matched training FLOPs JEM-ED dominates JEM-SGLD over the whole budget range on UCI-mixed, where every JEM-ED point has a higher C2ST-XGB than every JEM-SGLD point, and over the lower half of the range on the two CC18 suites (\Cref{fig:jem-cost}). At the top of the range on CC18, JEM-SGLD catches up (CC18-numerical, $K=20$) or overtakes (CC18-numerical at $K=40$, CC18-mixed at $K=20$).

\paragraph{Training cost at comparable quality}
Since $M$ barely matters for quality, the relevant comparison is JEM-ED at a small $M$ against JEM-SGLD at its default $K=20$. At $M=4$, JEM-ED uses $108\times$ fewer training FLOPs than JEM-SGLD on UCI-mixed ($4.4$ vs.\ $470$ TFLOP), $3.1\times$ fewer on CC18-numerical ($4.3$ vs.\ $13.1$) and $32\times$ fewer on CC18-mixed ($4.3$ vs.\ $136$), with higher accuracy on UCI-mixed ($0.854$ vs.\ $0.829$) but lower fidelity on CC18-numerical (C2ST-XGB $0.406$ vs.\ $0.430$) and CC18-mixed ($0.848$ vs.\ $0.879$). At the default $M=32$ the advantage shrinks to $16\times$ and $4.8\times$ on UCI-mixed and CC18-mixed, and reverses on CC18-numerical, where JEM-ED ($28.0$ TFLOP) costs $2.1\times$ more than JEM-SGLD ($13.1$ TFLOP) because an \emph{input}-gradient step costs two forward passes and a parameter update three, so $M{+}1$ parallel training passes are not cheaper than $K$ sequential input-gradient steps once $M$ exceeds roughly two thirds of $K$.

\paragraph{Wall-clock time}
FLOPs and wall-clock tell different stories. On CC18-numerical JEM-ED trains in $6$--$8$\,s for every $M$, about half the time of JEM-SGLD ($13$--$15$\,s), because its $M$ negatives are evaluated in parallel while SGLD's $K$ steps are sequential. On the two suites with categorical features, however, the parallelism is lost: JEM-ED's training time grows from $19$\,s at $M=1$ to $170$\,s at $M=32$ and $1{,}928$\,s at $M=128$ on UCI-mixed (JEM-SGLD: $32$--$62$\,s), and between $M=32$ and $M=128$ it grows $11\times$ for a $3.9\times$ increase in FLOPs. Since the same objective is flat in wall-clock on numerical data, this points to the current \emph{implementation} of the categorical perturbation rather than to the objective itself; for $M \le 4$ JEM-ED is faster than JEM-SGLD on every suite.

\paragraph{Quality of JEM-ED and JEM-SGLD}
\Cref{tab:main} compares the two JEMs at their best budgets ($M=2$ and $K=10$ on UCI-mixed, $M=64$ and $K=10$ on CC18). On UCI-mixed, JEM-ED is the better classifier on every predictive metric: accuracy increases from $0.827$ to $0.854$, AUROC from $0.836$ to $0.872$, and the conditional NLL and ECE drop from $0.481$ to $0.342$ and from $0.042$ to $0.015$. On CC18 the two are tied in accuracy ($0.829$ vs.\ $0.832$) and AUROC ($0.894$ vs.\ $0.893$), but JEM-SGLD is markedly better calibrated (ECE $0.077$ vs.\ $0.121$, NLL $0.493$ vs.\ $0.858$). On the generative side JEM-ED has the higher C2ST-XGB on both suites ($0.160$ vs.\ $0.104$ on UCI-mixed and $0.587$ vs.\ $0.483$ on CC18). In terms of the two scores, JEM-ED is ahead on both halves on UCI-mixed (prediction score $0.48$ vs.\ $0.37$, generation score $0.48$ vs.\ $0.43$); on CC18 JEM-SGLD is the better classifier ($0.54$ vs.\ $0.50$) and the two are equal as generators ($0.61$ vs.\ $0.60$). Since JEM-ED's quality barely depends on $M$, these comparisons hold at small $M$, at a fraction of JEM-SGLD's training cost.

\paragraph{ED on the marginal or on the full joint}
JEM-ED applies ED only to the marginal energy and trains the conditional $p_\theta(y \mid \vx)$ exactly with cross-entropy, whereas TabED-joint applies ED to the full joint, with the label as one more perturbed column, as in \citet{schroder2024heat}. Since both share every other setting, the difference between them isolates the effect of the JEM factorisation \eqref{eq:jem-loss}. At their best budgets (\Cref{tab:main}), the JEM factorisation gives the more accurate classifier (on UCI-mixed $0.854$ vs.\ $0.833$ and on CC18 $0.829$ vs.\ $0.820$), the better generator (generation score $0.48$ vs.\ $0.46$ and $0.61$ vs.\ $0.52$) and, most clearly, generated labels that depend more strongly on the features: MLE is $0.811$ vs.\ $0.797$ on UCI-mixed and $0.821$ vs.\ $0.740$ on CC18. Training the full joint with ED, on the other hand, gives the better-calibrated classifier on the small CC18 tables (Brier $0.257$ vs.\ $0.277$, ECE $0.068$ vs.\ $0.121$), so that its prediction score is slightly higher there. Note that the same pattern of weaker label dependence appears for the PC, the other joint model that treats the label as one more column (MLE $0.753$ on CC18).

\end{document}